\documentclass{SCIS2024}
\usepackage{amsmath,amsfonts}
\usepackage{algorithmic}
\usepackage{algorithm}
\usepackage{array}
\usepackage{subfig}
\usepackage{textcomp}
\usepackage{stfloats}
\usepackage{url}
\usepackage{verbatim}
\usepackage{graphicx}
\usepackage{cite}
\usepackage[colorlinks,hyperindex,breaklinks]{hyperref}
\usepackage{ulem}
\usepackage{algorithmic}
\usepackage{graphicx}
\usepackage{textcomp}
\usepackage{booktabs}
\usepackage{bigstrut}
\usepackage{epsfig}
\usepackage{pifont}
\usepackage{array}
\usepackage[table,xcdraw]{xcolor}
\usepackage{multirow}
\usepackage{xspace}
\usepackage{bbding}
\usepackage{pifont}
\usepackage{wasysym}
\usepackage{wrapfig}
\usepackage{utfsym}
\newcommand{\ie}{{\textit{i.e.}}\xspace}

\newcommand{\eg}{{\textit{e.g.}}\xspace}

\usepackage{arydshln}
\begin{document}
%\oa
%%%%%%%%%%%%%%%%%%%%%%%%%%%%%%%%%%%%%%%%%%%%%%%%%%%%%%%
%%% Authors do not modify the information below
%%% ×÷Õß²»ÐèÒªÐÞ¸Ä´Ë´¦ÐÅÏ¢
\ArticleType{RESEARCH PAPER}
%\SpecialTopic{}
%\luntan
\Year{2024}
\Month{}
\Vol{}
\No{}
\DOI{}
\ArtNo{}
\ReceiveDate{}
\ReviseDate{}
\AcceptDate{}
\OnlineDate{}

%%%%%%%%%%%%%%%%%%%%%%%%%%%%%%%%%%%%%%%%%%%%%%%%%%%%%%%

%%% title: ±êÌâ
%%%   \title{title}{title for citation}
\title{A Recover-then-Discriminate Framework for \\Robust Anomaly Detection}{Title keyword 5 for citation Title for citation Title for citation}

%%% Corresponding author: Í¨ÐÅ×÷Õß
%%%   \author[number]{Full name}{{email@xxx.com}}
%%% General author: Ò»°ã×÷Õß
%%%   \author[number]{Full name}{}
\author[1]{Peng XING}{}
\author[2]{Dong ZHANG}{}
\author[1]{Jinhui TANG}{}
\author[1]{Zechao LI}{{zechao.li@njust.edu.cn}}

%%% Author information for page head. Ò³Ã¼ÖÐµÄ×÷ÕßÐÅÏ¢
\AuthorMark{Peng Xing}

%%% Authors for citation. Ê×Ò³ÒýÓÃÖÐµÄ×÷ÕßÐÅÏ¢
\AuthorCitation{Author A, Author B, Author C, et al}

%%% Authors' contribution. Í¬µÈ¹±Ï×
% \contributions{This work was supported by the National Natural Science Foundation of China (Grant No. U21B2043).}
% \Foundation{This work was supported by the National Natural Science Foundation of China (Grant No. U21B2043).}
%%% Address. µØÖ·
%%%   \address[number]{Affiliation, City {\rm Postcode}, Country}
\address[1]{School of Computer Science and Engineering, Nanjing University of Science and Technology, Jiangsu,  {\rm 210094}, China}
\address[2]{Department of Electronic and Computer Engineering, The Hong Kong University of Science and Technology, Hong Kong, {\rm 999077}, China. }
% \address[3]{Affiliation, City {\rm 000000}, Country}

%%% Abstract. ÕªÒª
\abstract{Anomaly detection (AD) has been extensively studied and applied in a wide range of scenarios in the recent past. However, there are still gaps between achieved and desirable levels of recognition accuracy for making AD for practical applications. In this paper, we start from an insightful analysis of two types of fundamental yet representative failure cases in the baseline model, and reveal reasons that hinder current AD methods from achieving a higher recognition accuracy. Specifically, by Case-1, we found that the main reasons detrimental to current AD methods is that the inputs to the recovery model contain a large number of detailed features to be recovered, which leads to the normal/abnormal area has-not/has been recovered into its original state. By Case-2, we surprisingly found that the abnormal area that cannot be recognized in image-level representations can be easily recognized in the feature-level representation. Based on the above observations, we propose a novel Recover-then-Discriminate (ReDi) framework for AD. ReDi takes a self-generated feature map and a selected prompted image as explicit input information to solve problems in case-1. Concurrently, a feature-level discriminative network is proposed to enhance abnormal differences between the recovered representation and the input representation. Extensive experimental results on two popular yet challenging AD datasets validate that ReDi achieves the new state-of-the-art accuracy.}
% \input{../TIIsections/1abs}
%%% Keywords. ¹Ø¼ü´Ê

\maketitle

%%%%%%%%%%%%%%%%%%%%%%%%%%%%%%%%%%%%%%%%%%%%%%%%%%%%%%%
%%% The main text. ÕýÎÄ²¿·Ö
%%%%%%%%%%%%%%%%%%%%%%%%%%%%%%%%%%%%%%%%%%%%%%%%%%%%%%%
\section{Introduction}
% ----------------------------------------
Image anomaly detection (AD) aims to detect anomalous areas, which deviate from expected patterns, within a given image~\cite{deng2022anomaly,tao2022deep,Zhang_1_2023_CVPR,hou2022collaborative,xing2023adps}.
In the last few years, this task has been extensively studied and applied in various practical applications, \eg, medical diagnosis~\cite{Xiang_2023_CVPR}, scene surveillance~\cite{xue2020anomaly}, and industrial inspection~\cite{cheng2023atomgan,zhan2021spar,zavrtanik2021draem,Roth_1_2022_CVPR,Ding_2022_CVPR,Yao_2023_CVPR,Gu_2023_ICCV}. 
Although AD has achieved impressive results, this task remains a challenging research topic. In particular, sometimes the abnormal area only occupies a small part of the entire image~\cite{bergmann2019mvtec,bovzivc2021mixed}. %Besides, since it is difficult to collect enough images containing various abnormal patterns, most anomaly detection methods only use normal samples for training~\cite{bergmann2020uninformed,deng2022anomaly,defard2021padim}.
Besides, it is difficult to collect enough images containing various abnormal patterns. Therefore, most anomaly detection models only use normal samples for model training~\cite{bergmann2020uninformed,deng2022anomaly,defard2021padim} and cannot be based on supervised detection~\cite{DBLP:journals/pami/LiSZT22,zhang2020causal} methods, which are called unsupervised anomaly detection or self-supervised anomaly detection methods.
% ----------------------------------------
\begin{figure*}
\centering
\includegraphics[width=.65\linewidth]{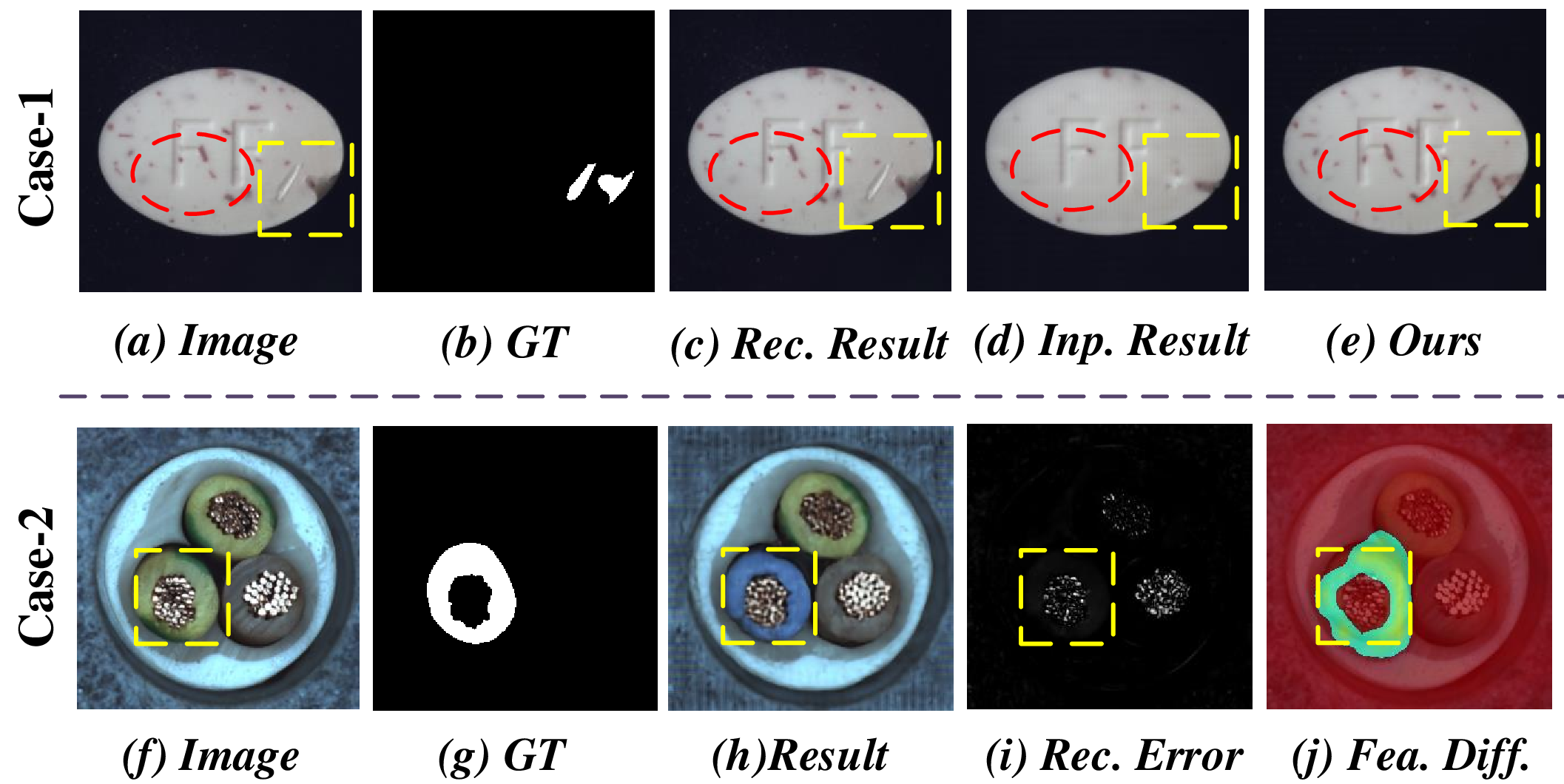}
\caption{%
{The key motivations of this paper. In \textbf{Case-1}, we found that the main reason detrimental to current approaches for anomaly detection is that the normal area has not been recovered to its original state, and the abnormal area has been recovered to its original state.}
In \textbf{Case-2}, we found that the abnormal area that cannot be recognized in the image-level representation can be easily recognized in the feature-level representation. The red/yellow dotted boxes highlight normal/abnormal areas.``\textit{GT}'', ``\textit{Rec.}'', ``\textit{Inp.}'', and ``\textit{Fea. Diff.}'' are ``Ground-Truth'', ``Reconstructed'', ``Inpainting'', and ``Feature-level Differences'', respectively.}
\label{fig:figure1} 
\end{figure*}
% ----------------------------------------

Recovery-based methods -- image reconstruction and image inpainting -- have found wide applications in AD~\cite{gong2019memorizing,bergmann2018improving, ZavrtanikKS21, hou2021divide}. These methods are designed to learn the normal image recovery process, and subsequently detect anomalous regions based on large recovery errors~\cite{bergmann2018improving,bergmann2019mvtec}. However, due to significant disparities between the distribution of abnormal and normal images, the recognition accuracy of such methods is usually unsatisfactory~\cite{gong2019memorizing}, falling short of the actual needs~\cite{gong2019memorizing,fei2020attribute}. This problem highlights the urgent need for improved accuracy in AD methods, which can enhance the effectiveness of recovery-based approaches~\cite{you2022unified,gong2019memorizing}. In this work, to explore fundamental factors that impede the current AD methods from achieving superior recognition accuracy, as illustrated in Figure~\ref{fig:figure1}, we first start from the analysis of two types of fundamental yet representative failure cases in the baseline model~\cite{bergmann2018improving}. 

{
As illustrated in \textbf{Case-1} of Figure~\ref{fig:figure1}, we show two results from the recovery-based methods of image reconstruction~\cite{bergmann2018improving} and image inpainting~\cite{ZavrtanikKS21} in AD. 
%
% However, in such methods, the original image information is already fully present in the input image (a), thus making the restoration easy and all images can be accurately recovered~\cite{hou2021divide,salehi2020puzzle,zavrtanik2021reconstruction}. 
%
However, in such methods, they directly use the original images for their input, thus making the recovery and restoration work easy and ensuring that all images are accurately recovered.
As an example, due to the lack of anomalous training samples, models trained via image reconstruction methods only utilizing normal samples can be subject to ``identical shortcut''~\cite{you2022unified} learning, and hence may not be suitable for accurate recovery of both normal and abnormal regions in the input image, as evidenced by the yellow bound-boxes.} 
Image inpainting methods rely on contextual information to infer eliminated areas, but such information may still contain the original image information, thereby affecting the recovery of eliminated areas~\cite{guo2021image,tao2022unsupervised}. Furthermore, eliminating structural and detailed information during the process may result in the loss of vital details in the recovery results~\cite{suvorov2022resolution,lugmayr2022repaint}, as marked by the red dotted boxes.

{
As illustrated in \textbf{Case-2} of Figure~\ref{fig:figure1}, 
We observe that the recovery error \textit{(i)} obtained from the original image \textit{(f)} and the recovered image \textit{(h)} in RGB space has a large deviation from Ground-Truth \textit{(g)} leading to a detection failure, even though \textit{(f)} and \textit{(h)} have a large visual difference (color difference) in the anomalous region (denoted by \textit{(g)}).}
This phenomenon can be expounded upon from the following two perspectives. 
The recovery model cannot guarantee that the recovery error of normal regions in pixel dimension is extremely low. 
For example, as shown in Figure~\ref{fig:figure1}~\textit{(i)},  the recovery error in \textbf{Case-2} focuses on the normal regions in the lower right corner of the image. From a visual point of view, the differences in the recovery of these normal regions may be small, but in reality, the pixel-level errors are large~\cite{hou2021divide,zhang2021self}.
%
% we can observe a notable phenomenon whereby the regions exhibiting abnormality that are indiscernible to the AD model in the image-level representation can be detected with ease in the feature-level representation. This phenomenon can be expounded upon from the following two perspectives: the recovery model cannot guarantee that the recovery error of normal regions in pixel dimension is extremely low. For example, as shown in Figure~\ref{fig:figure1}~(i),  the recovery error in \textbf{Case-2} focuses on the normal regions in the lower right corner of the image. From a visual point of view, the differences in the recovery of these normal regions may be small, but in reality, the pixel-level errors are large~\cite{hou2021divide,zhang2021self}.
%
%the recovery error is rendered ineffectual in detecting anomalies, even when there exist significant visual discrepancies between the anomalous and original areas of the recovered images~\cite{hou2021divide,zhang2021self}. 
%
{
In addition, the misalignment of the recovery image with the original image at the boundary is a potential factor leading to the false detection of anomalous regions~\cite{li2022mat, ZavrtanikKS21}. 
In other words, if the recovery model is based on the inpainting method, the images recovered from the masked region may be varied and not necessarily the same as the original input image, but they may all be normal.}
Since normal images have diversity, although the recovered images are conveniently slightly misaligned with the original images generating recovery errors, as long as they are semantically similar in high-dimension space they should be recognized as normal classes. 
This leads to recovery error failure in RGB space, \ie, regions with recovery errors may be normal.

To address these two failure cases, a novel AD framework -- \textbf{Re}cover then \textbf{Di}scriminate (\textbf{ReDi}) -- is proposed, which includes a Recover Network to overcome the ``identical shortcut" challenge of image recovery and a Discriminate Network to discriminate anomalies in feature space.
{
In the Recover Network, we propose to use the self-generated map with image prompt (HIP) method to address problems in {Case-1}. \textit{Firstly}, HIP discards the original image and replaces it with a self-generated map (\ie, HOG feature) as input. The self-generated map can provide the low-level information needed for recovery without exposing the detailed information of the original image. \textit{Secondly}, HIP introduces a normal image with high similarity to the original image as the prompt image to provide semantic information to guide image recovery, which ensures the accurate recovery of normal regions. 
Co-guidance of low-level features and normal semantic features ensures ideal recovery results, with accurate recovery of normal regions and large errors in the recovery of abnormal regions.
In the Discriminate Network, we propose a reference branch and a recovery branch to extract multi-scale features of the original image and recovery image, respectively, and compare the differences in the feature space to identify anomalies, such that the finding in {Case-2} can be exploited.
To reduce the difference extracted by the two branches for the normal region, the recovery feature extracted from the recovery branch is required to be the same as the reference feature extracted from the reference branch by the pre-trained model. To enhance the difference for the abnormal region, the recovery branch further introduces a feature recovery block to filter anomalous features, by feature aggregation and transposed convolution.
Besides, we propose a self-correlation loss to ensure the consistency of normal features in two branches. By the ``self-correlation'', we mean the correlation relationship between features at different locations. The self-correlation loss constrains this relationship to be the same in the reference feature map and the recovery feature map. Extensive experimental results on two popular yet challenging AD datasets validate that \textit{ReDi} achieves the state-of-the-art recognition accuracy.}
The main contributions are summarized as follows:

\begin{itemize}
% \item We carefully analyze the shortcomings of current recovery methods and propose a new recovery network HIP and discriminative network to mitigate the shortcomings. 
\item A self-generated feature map and a selected prompted image are used as explicit input information to solve potential fundamental problems in AD.   
\item A feature-level discriminative network is proposed to enhance abnormal differences between the recovered representation and the input representation. 
\item {A self-correlation loss is proposed to further constrain the recovery features of normal samples to align with the reference features.}
% \item Our proposed ReDi achieves the state-of-the-art accuracy on two popular yet challenging AD benchmark datasets.
\end{itemize}

{
The paper is structured as follows. First, we present related work and review anomaly detection methods in Section~\ref{sec2}. 
In Section~\ref{sec3}, we introduce the proposed ReDi framework in detail, including the HIP-based Recovery Network and the feature space-based Discriminate Net. 
In Section~\ref{sec4}, we provide the state-of-the-art performance of the proposed ReDi on the anomaly detection benchmark datasets and compare it with recent approaches. Finally, we conclude in Section~\ref{sec5}.
}
% ----------------------------------------
\section{Related Work}\label{sec2}
\subsection{{Hand-crafted Image Descriptors}}

{
Early work~\cite{viola2004robust,dalal2005histograms,bay2006surf} explore some handcrafted image descriptors to describe the low-level information of an image. The edge detection operator is one of the simplest image descriptors that describe the mutation places of information such as gray scale or structure. 
It mainly includes Sobel~\cite{duda1973pattern}, Robert operator~\cite{roberts1963machine} based on first-order derivative detection, Laplace operator with second-order derivative detection, and canny~\cite{Canny86a}, an optimization operator for edge detection derived by satisfying certain constraints.  
%Canny86a,duda1973pattern,roberts1963machine
In addition, some more advanced image descriptors such as Haar~\cite{viola2004robust}, HOG~\cite{dalal2005histograms}, SIFT~\cite{lowe2004distinctive}, SURF~\cite{bay2006surf} have been used in most computer vision applications. Haar~\cite{viola2004robust} features are based on the Haar wavelet transform, which is usually used to detect local features in an image, such as edges, lines, and corners.
SIFT~\cite{lowe2004distinctive} is able to extract the key points in an image at different scales and rotation angles and compute the local feature descriptors of these key points for image matching and recognition.
% }%
%
HOG~\cite{dalal2005histograms} is a traditional feature descriptor that describes the distribution of gradient directions within a local region. The detailed level of low-level information in the HOG image depends on the number of directions ($bin$) and the size ($size$) of the local region. HOG features are widely used in applications such as pedestrian detection \cite{wang2009hog}, face recognition~\cite{pang2011efficient}, and classification~\cite{zhang2021hog,cao2011linear} due to their excellent spatial and optical invariance. For example, MaskFeat~\cite{wei2022masked} proposes a self-supervised task to predict the HOG features of the original image, obtaining an advanced self-supervised pre-trained model. These work show that HOG features are related to image detail features~\cite{zhang2023augmented}.
In the proposed \textit{ReDi}, the HOG feature guarantees the detailed outline and low-level information of input without exposing the original anomalous information. }

\subsection{Prompt Learning} Prompt learning has been widely usually used in NLP and CV interaction tasks, such as pre-trained models and visual question answer~\cite{liu2023pre}. CLIP~\cite{radford2021learning}, CoOp~\cite{zhou2022learning}, and other work~\cite{zhou2022conditional} employ textual prompts achieving powerful performance. For example, CLIP constructs ``$\mathrm{a\ photo\ of\ a\ [CLS]}$ " as the textual prompt for the corresponding image, where ``$\mathrm{CLS}$'' generally refers to the actual class name (e.g., $\mathrm{a\ photo\ of\ a\ cat}$ ). CLIP has shown strong performance in VQA task \cite{shen2021much} and image classification task. Recently, DenseCLIP~\cite{rao2022denseclip} using textual prompts has shown significant performance improvement in semantic segmentation task. It utilizes similar textual prompts and extracts text features with image features to achieve supervised semantic segmentation. We extend prompt learning by introducing image prompt into the image recovery for AD. The purpose is to ensure that the contextual information of the recovery process is normal semantic information.

\subsection{Anomaly Detection (AD)} 
{
AD focuses on the detection of anomalous images and segmentation of anomalous regions by training exclusively with normal samples~\cite{Gu_2023_ICCV}. }

{
\emph{Distribution-based} approach relies on pre-trained models to extract the features of normal samples and then model feature distribution~\cite{DBLP:journals/corr/abs-2111-07677,defard2021padim,DBLP:conf/cvpr/RothPZSBG22,tang10342826}.
In the inference stage, features that do not fall within the boundary of the normal feature distribution are detected as anomalies, as in the earlier work DSVDD~\cite{DBLP:conf/icml/RuffGDSVBMK18}. 
However, directly modeling the global image feature results in losing detailed information. Therefore, SPADE~\cite{cohen2020sub} and PaDim~\cite{defard2021padim} attempted to extract patch features.
SPADE discriminates anomalies by comparing the presence of $k$ with similar normal patches' features to be detected.
PaDim further employs a Gaussian mixture model to model the relationship between patch and context, which is not effective enough for complex contextual anomaly scenarios. 
Another class of feature modeling approaches employs a method known as normalized flow~\cite{DBLP:conf/icml/RezendeM15}, which constructs complex distributions through a series of reversible transformations. 
Generally, these methods establish an invertible mapping relationship between prior features and Gaussian distributions~\cite{DBLP:journals/corr/abs-2111-07677,DBLP:conf/wacv/RudolphWR21,DBLP:journals/corr/abs-2303-02595,gudovskiy2022cflow,DBLP:conf/wacv/RudolphWRW22}. 
The anomalous features deviate from the Gaussian distribution after normalized flow, while the normal features conform to the Gaussian distribution. For example, CS-Flow~\cite{DBLP:conf/wacv/RudolphWRW22} introduces cross-convolution and multi-scale features to identify anomalies, and CFlow-AD~\cite{gudovskiy2022cflow} combines location coding to improve anomaly detection performance. 
However, these methods are not effective for high-accuracy anomaly segmentation.
}

{
\emph{Knowledge distillation-based} approaches rely on feature distillation to train one or more uninformed student models that learn normal sample expressions from the teacher model~\cite{zhang2021self}. U-Std~\cite{bergmann2020uninformed} uses multiple student networks and compares the differences in a high-dimensional space. However, it is limited by the scale of the features that cannot solve the anomaly segmentation challenge. MKDAD~\cite{salehi2021multiresolution} uses the VGG network~\cite{simonyan2014very} as the backbone and adopts a multi-scale feature comparison. However, the shallow student network weakens the representation capability. STPM~\cite{wang2021student} and RSTPM~\cite{yamada2021reconstruction} focus on the selection of pre-trained models and the number of feature layers for comparison. RD~\cite{deng2022anomaly} proposed the reverse distillation paradigm to further enhance the anomaly detection capability. In summary, the distillation approaches have the advantage of leveraging prior knowledge of the pre-trained model. The drawback is that the student network can extract anomalous information directly, leading to the possibility that the features extracted by both networks are identical.}

% Frequently studied anomaly detection methods include \emph{recovery-based}, \emph{distribution-based}, and \emph{knowledge distillation-based} anomaly detection methods.

{
\emph{Recovery-based} methods expect the models to reconstruct normal regions with low recovery error and abnormal regions with large recovery error, which directly compares pixel-level reconstruction errors can be realized for anomaly detection and anomaly segmentation.
In previous work, AE-SSIM~\cite{bergmann2018improving} was proposed to train an autoencoder by self-supervised training on normal samples, with the aim of learning only the recovery of normal samples. 
However, the powerful generalization ability of the autoencoder allows anomalous regions to be easily reconstructed~\cite{fei2020attribute,you2022unified}, called ``identical shortcut". 
There are currently two types of methods that expect to alleviate ``identical shortcut".
One class of methods expects to introduce limiting structures (\eg, memory banks) in the middle of the autoencoder, constraining anomalous feature inputs to the decoder, such as MemAE \cite{gong2019memorizing}, DAAD~\cite{hou2021divide}, PMB-AE~\cite{xing2023visual}, and Con-Mem~\cite{wang2021cognitive}.
However, it corrupts the expressiveness of the autoencoder and leads to mediocre anomaly detection.
Another type explores different self-supervised tasks to train the autoencoder, such as image inpainting~\cite{ZavrtanikKS21}, rotated image reconstruction ARFAD~\cite{fei2020attribute}, and puzzle task reconstruction Puzzle-AE~\cite{salehi2020puzzle}. 
However, the encoders of these self-supervised methods receive the original information of the recovery results, leading to still poor practical results. 
In this work, the proposed \textit{ReDi} first trains a recover network to avoid recovery shortcuts by introducing HOG and prompt images for the recovery task, and then trains a discriminate network to leverage the prior knowledge of the pre-trained model, thereby identifying the disparities between the original and recovery images.
}
\section{Our Approach}\label{sec3}
%review the commonly used image reconstruction and image inpainting methods (\textit{Ref. Sec~\ref{sec:3.A}}). Then, we 
{In this section, we first introduce step by step the proposed \textbf{Re}cover then \textbf{Di}scriminate framework, henceforth referred to as \textit{ReDi}. \textit{ReDi} is a two-stage anomaly detection approach, in which the first stage trains an image recovery network and the second stage trains a feature discriminate network.} The recover network (\textit{Ref. Sec~\ref{sec:3.1}}) utilizes structural and semantic information to recover images, thereby addressing the challenge presented in \textbf{Case-1} (Figure \ref{fig:figure1}). The discriminate network (\textit{Ref. Sec~\ref{sec:3.2}}), by differentiating between the recovered and original images in the feature space, aims to detect anomalous regions and tackle the challenges presented in \textbf{Case-2} (Figure \ref{fig:figure1}). Importantly, \textit{ReDi} operates as an unsupervised anomaly detection scheme, eliminating the need for human-provided prior information to generate simulated anomaly samples.

% ----------------------------------
\begin{figure*}[t]
\centering
\includegraphics[width=0.99\linewidth]{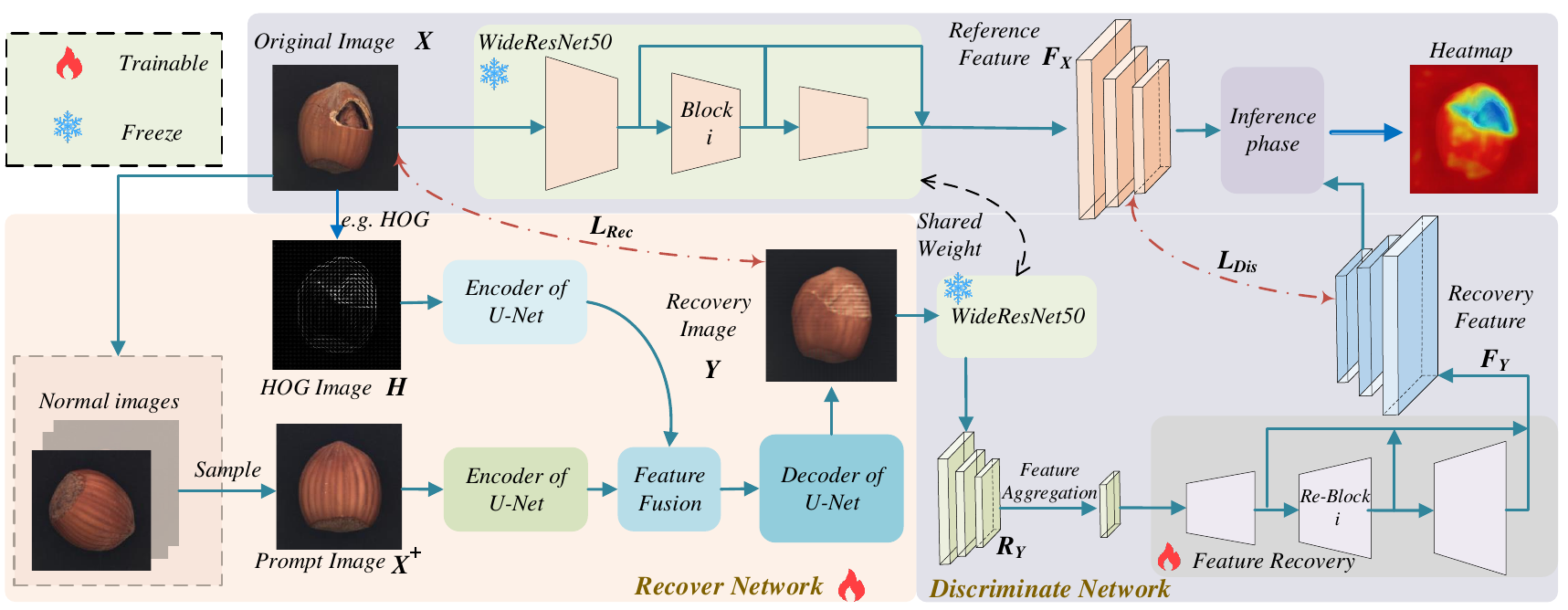}
\caption{{Overview of the proposed \textit{ReDi} framework. 
Given an input image $X$, the self-generated map $H$ (\eg, HOG) is first extracted and the prompt image is sampled from the set of normal image $X^+$. Then, the recovery network relies on the self-generated graph and prompt image to generate the recovery image $Y$.
 The reference branch extracts features of $X$ as the reference feature $F_X$.  The recovery branch first extracts the feature $R_Y$ of the recovery image $Y$, and then inputs it to the Feature Recovery Block to generate recovery feature $F_Y$. The anomalous regions are inferred by $F_X$ and $F_Y$ in the inference phase. $L_{Rec}$ and $L_{Dis}$ denote loss functions of two networks respectively.} }
\label{overview1} 
\vspace{-0.2cm}
\end{figure*}
\subsection{Recover Network}
\label{sec:3.1}
In popularly used image reconstruction and image inpainting methods~\cite{bergmann2018improving,zavrtanik2021reconstruction}, the original image $X$ is used as the input to the corresponding model. 
$X$ contains all the features (high and low dimensional information) needed to recover the image, which leads to the risk that Recover Network can recover precisely anomalous regions. 
Towards this end, we propose an image prompt-based recovery method in Recover Network called HOG with Image Prompt (HIP). {HIP utilizes the self-generated feature map of the original image instead of the original image as input, allowing the model to never capture the original detailed information leading to simple recovery shortcuts. Meanwhile, it introduces the normal prompt image to ensure the recovery of normal areas.}  Specifically, HIP utilizes HOG images to provide shape and appearance attributes (low-level information) while utilizing selected normal images as prompt images, and the recovery model is required to recover normal images through self-generated maps and high-dimensional normal semantics.
Empirically, traditional features (\eg, HOG and Canny~\cite{Canny86a}) that capture the local shape without exposing the original detailed information as the self-generated maps are effective. Based on our experiments (\textit{Ref. Section~\ref{431}}), HOG is the optimal choice.

{
As shown in the Recover Network in Figure~\ref{overview1}, the input of HIP contains two parts: the HOG image $H$ and prompt image $X^+$. The prompt image $X^+$ is not the same as the original input $X$. 
First, we will describe the prompt image in detail, and then we will describe how the HIP works.
Because of the diversity among normal samples, randomly sampling a normal sample as the prompt image may not be effective in guiding accurate recovery of the HOG image of the normal sample.
 To ensure that $X^+$ provides valid normal semantic information, we define $X^+$ as a normal image similar to $X$. 
 To sample the normal image $X^+$ that is most similar to the original image $X$ as possible, we utilize a pre-trained model (WideResNet50~\cite{DBLP:conf/bmvc/ZagoruykoK16}) to extract 2048-dimensional features as a similarity metric.
 Next, we compute the cosine similarity of features between normal samples and $X$ and use the sample corresponding to the highest similarity as the prompt image $X^+$.
In both the inference and training phases, the candidate set of prompt images are all normal samples.
}

On HIP, first, we extract the multi-scale features of $H$ and $X^+$ by two independent feature extractors (We use the encoder of the UNet with each extractor dimension changed to half of the original.), respectively. This process can be expressed as:
\begin{equation}
	T_H = \mathcal{N}_1(H),
        T_{X^+} =\mathcal{N}_2(X^+),
\end{equation}
where $\mathcal{N}_1(\cdot)$ and $\mathcal{N}_2(\cdot)$ represent two independent feature extractors. Then, $T_H$ and $T_{X^+}$ represent the corresponding features, which are concatenated along the channel dimension by Feature Fusion Block, which is formulated as:
 \begin{equation}
 	T_C = cat(T_H,T_{X^+}),
 \end{equation}
where $T_C$ represents the output of Feature Fusion Block. Finally, we recover the image by using the fused features through a decoder. This process can be expressed as:
 \begin{equation}
 	Y = Dec(T_C),
 \end{equation}
 where $Y$ denotes the recovery image and $Dec(\cdot)$ denotes the decoder of the UNet structure. To train Recover Network, the commonly used $L_2$ loss and multi-scale gradient magnitude similarity loss $L_M$~\cite{zavrtanik2021reconstruction,7952357} are introduced.
%loss function is introduced. However, $L_2$ loss constrains the relationship between $X$ and 
%$Y$ corresponding pixel pairs and does not explore the connection between individual image pixels. Therefore, a multi-scale gradient magnitude similarity (MSGMS) loss \cite{zavrtanik2021reconstruction} (Appendix \ref{appendixB}) that measures the structural difference between $X$ and $Y$ is introduced. 
The overall loss $L_{Rec}$ of the Recover Network is expressed as:
\begin{equation}
	{L_{Rec}}(X,{\rm{Y}}) = {L_2}(X,Y) + {\lambda _M}{L_M}(X,Y),
\end{equation}
where $\lambda _M$ is a weight for loss balancing.

%-------------------------------------------------------------------------
\subsection{Discriminate Network}
\label{sec:3.2}
A Discriminate Network is proposed to identify the differences between the recovery image and the original image in the feature space utilizing prior knowledge of the pre-trained model. As shown in Discriminate Network in Figure~\ref{overview1}, Discriminate Network contains a reference branch and a recovery branch. The reference branch is a pre-trained WideResNet-50 network (as \textit{Feature Extractor $3$}) that is responsible for extracting multi-scale features of the original image as reference feature $F_X=\{F_X^1, F_X^2,..., F_X^n\}$. In the recovery branch, the multi-scale features $R_Y$ of recovery image $Y$ are first extracted by the same pre-trained model. 
{Since the recovery image $Y$ and the original image $X$ may have subtle recovery errors in the normal region, there may still be subtle differences between $R_Y$ and $F_X$ in the feature space.}
Towards this end, this paper introduces a learnable Feature Recovery Block, which aims to further optimize the features $R_Y$ by learning the representation of reference branch for normal regions.
{
We follow RD~\cite{deng2022anomaly} by first aggtegating $R_Y$ to a smaller dimension and then generating recovery feature $F_Y =\{F_Y^1, F_Y^2,..., F_Y^n\}$ with the same dimension as $F_X$ by multiple transposed convolution layers.} Formally, 
%feature compression is compresses the features $R_Y$ extracted from the pre-trained model of the recovery branch, and then recovering them to recovery feature $F_Y$ by transposed convolution layers. In general,
%
$R_Y$ is a multi-scale feature map, $R_Y=\{R_Y^1, R_Y^2,..., R_Y^n\}$, where $ R_Y^{n-1} \in \mathcal{R}^{4h\times 4w \times 4c},  R_Y^n \in \mathcal{R}^{ 2h\times 2w \times 2c}$. Feature aggregation is to down-sample each item in $R_Y$ to a smaller scale, such as $h\times w\times c$.
{
In this paper, it is implemented by $3\times3$ convolution layers and the ReLU activation function. Then, we can obtain the aggregation results by:}
% In this paper, it is implemented by $3\times3$ convolution layers ($Conv(\cdot)$) and ReLU activation function ($\sigma(\cdot)$).
	\begin{equation}
	\begin{array}{l}
		% RF_Y^1 = {\sigma(Conv}( \cdot  \cdot  \cdot ({\sigma(Conv}(R_Y^1))))), RF_Y^1 \in {\mathbb{R}^{h \times w \times c}},\\
		% ...,\\
		% RF_Y^n = {\sigma(Conv}(R_Y^n),RF_Y^n \in {\mathbb{R}^{h \times w \times c}}.\\
		Output = Cat(RF_Y^1,...,RF_Y^n),Output \in {\mathbb{R}^{h \times w \times nc}},
	\end{array}
\end{equation}
$Cat(\cdot)$ the multiple feature maps concatenation along the channel dimension, $Output$ is taken as the input of the Feature Recovery Block, which is generated by the recovery feature $F_Y$. The size of $F_Y$ is the same as $R_Y$. The purpose of feature compression is to further filter anomalous patterns in $R_Y$~\cite{deng2022anomaly}. During the training phase, we constrain the consistency of the recovery features and the reference features to optimize the Feature Recovery Block. 
%this paper follows RD \cite{deng2022anomaly},  proposes a two branches (reference and recovery branches) Discriminate Net. Different from RD, we take the recovery image obtained in Recover Net as the input of the recovery branch and the original image as the input of the reference branch. Besides, we propose a novel self-correlation loss function to explore the relationship between image pixels. 
%
% In the reference branch, the multiscale features of $X$ are extracted as reference features $F_X=\{F_X^1, F_X^2,..., F_X^n\}$ utilizing the pre-trained model. In the recovery branch, the multiscale features $R_Y$ of $Y$ are first extracted utilizing the same pre-trained model. 
% Since the generation of recovery image is ill-posed problem, the recovery image $Y$ and the original image $X$ may have a fine recovery error in the normal region. The normal regions with subtle recovery errors may be misclassified as abnormal regions. In particular, the differences in the high-dimensional feature space may be amplified. For this reason, this paper proposes the learnable Feature Reconstruction Block, which aims to further recover the features $R_Y$ increasing the recovery tolerance of the model for normal regions. The input of the Feature Reconstruction Block is $R_Y$ after feature compression. Then, the recovered features $F_Y$ of the same size as $F_Y =\{F_Y^1, F_Y^2,..., F_Y^n\}$ are generated by transposed convolutional. Detailed description is provided in Appendix \ref{appendixC}.
To constrain the $F_Y$ alignment to $F_X$, the cosine similarity loss ${L_D} $ is introduced. It can be formulated as:
\begin{equation}
	{L_D}(F_X,F_Y) = \frac{1}{n}\sum\limits_i^n {(1 - \frac{{F_X^i \cdot F_Y^i}}{{\left\| {F_X^i} \right\| \times \left\| {F_Y^i} \right\|}}} ) \label{eq4}
\end{equation}
Besides, we also propose a self-correlation loss ${L_S}$. 
By ``correlation", it means the correlation between multiple features, and the correlation between two features is expressed as the inner product between feature vectors.
By ``Self", refers to the correlation of features at different locations in the own feature map.
First, we calculate the self-correlation $G_X=\{g_X^1, g_X^2,... , g_X^n\}$ of $F_X$ and the self-correlation $G_Y=\{g_Y^1, g_Y^2,... , g_Y^n\}$ of $F_Y$. 
\begin{equation}
	g_X^i = \hat{(F_X^i)} ^T\cdot \hat{F_X^i},g_Y^i = \hat{(F_Y^i)}^T \cdot \hat{F_Y^i}
\end{equation}
where $\hat{F_X^i}$,$\hat{F_Y^i} \in \mathbb{R}^{(c\times hw)}$ are $F_X^i,F_Y^i \in \mathbb{R}^{(c\times h\times w)}$ be normalized by softmax and be reshaped. Then, self-correlation loss ${L_S}$ constrain the distance of $G_X$ and $G_Y$ by $L_2$ loss.
\begin{equation}
	{L_S}({F_X},{F_Y}) = \frac{1}{n}\sum\limits_i^n {{{\left\| {g_X^i - g_Y^i} \right\|}_2}} 
\end{equation}
The overall loss $L_{Dis}$ of the Discriminate Network is expressed as
\begin{equation}
	{L_{Dis}}(F_X,{\rm{F_Y}}) = {L_D}(F_X,F_Y) + {\lambda _S}{L_S}({F_X},{F_Y}) ,
\end{equation}
where $\lambda _S$ is the loss weight.

\subsection{Model Inference} 
In inference, the regions with different distributions of $F_X$ and $F_Y$ are detected as abnormal regions. We use the cosine similarity function to measure whether the distributions are identical or not. Therefore, this process can be formulated as:
\begin{equation}
{D_{(i,j)}} = \sum\limits_k^n {(1 - \frac{{{{(F_X^k)}_{(i,j)}} \cdot {{(F_Y^k)}_{(i,j)}}}}{{\left\| {{{(F_X^k)}_{(i,j)}}} \right\| \times \left\| {{{(F_Y^k)}_{(i,j)}}} \right\|}}}),
\end{equation}
where $D_{(i,j)}$ represents the anomaly score of the coordinate$(i,j)$.  The overall anomaly score of image $X$ is the maximum value of $D_{(i,j)}, i \in [1,h], j\in [1,w]$.
% ----------------------------------------
\section{Experiments}\label{sec4}

%------------------------------------------------------------------------

\subsection{Experimental Settings}
\subsubsection{Datasets.} Our experiments are conducted on two challenging datasets, MVTec-AD~\cite{bergmann2019mvtec} and KolektorSDD2~\cite{bovzivc2021mixed}. \textbf{MVTec-AD}~\cite{bergmann2019mvtec} is a comprehensive and anomaly type-rich industrial anomaly detection benchmark dataset with 15 categories, including 5 categories for textured images (Textures) and 10 categories for non-textured images (Objects). Following recent work on anomaly detection \cite{hou2021divide,bergmann2020uninformed,li2021cutpaste}, we evaluate the anomaly detection model in 15 sub-categories. \textbf{KolektorSDD2}~\cite{bovzivc2021mixed} is a surface defect detection dataset containing more than 3000 images, which contains tiny scratches, dots, surface defects, etc.

\subsubsection{Evaluation Metrics.} Following prior works \cite{hou2021divide,bergmann2020uninformed,li2021cutpaste,cohen2020sub}, the Area Under the Receiver Operating Curve (AUROC) is used as the evaluation metric for anomaly detection and anomaly segmentation. Besides, on the anomaly segmentation, due to the extreme imbalance between anomalous pixels and normal pixels, AUROC may present an exaggerated view of performance~\cite{zou2022spot,tao2022deep,zavrtanik2021draem}. This paper follows ~\cite{tao2022deep,zavrtanik2021draem} to introduce Average precision (AP), which is a more reasonable measure of anomaly segmentation capability.  $\mathcal{AUROC}_{det}$, $\mathcal{AUROC}_{seg}$, and $\mathcal{AP}_{seg}$ denote AUROC for anomaly detection, AUROC for anomaly segmentation, and AP for anomaly segmentation, respectively.

\subsubsection{Implementation Details} The size of the input image is set to a specific resolution of $256 \times 256$. The HOG image is generated by the setting with $bin$ of $9$ and $size$ of $8 \times 8$. The recover Network is trained with 500 epochs. The learning rate is set to $5 \times 10^{\textnormal{-}4}$ initially, and dropped by $0.5$ after $200/400$ epochs. The Discriminate Network is trained with $300$ epochs. The learning rate is $5 \times 10^{\textnormal{-}3}$ initially, and dropped by $0.5$ after $100 / 200$ epochs. The batch size is set to 32, and the AdamW optimizer~\cite{loshchilov2017decoupled} is employed. $\lambda_M$ and $\lambda_S$ are set to $1$.

\subsubsection{Baselines} Our approach is compared with advanced baseline methods including: SPADE  ~\cite{cohen2020sub}, U-Std~\cite{bergmann2020uninformed}, MKDAD~\cite{salehi2021multiresolution}, DAAD+ \cite{hou2021divide}, RIAD~ \cite{zavrtanik2021reconstruction}, MAD~\cite{rippel2021modeling}, Cutpaste~    \cite{li2021cutpaste}, SGSF~\cite{xing2022self}, MF~\cite{wu2021learning}, PaDim~\cite{defard2021padim}, DRAEM~\cite{zavrtanik2021draem}, AnoSeg~\cite{DBLP:journals/corr/abs-2110-03396}, RD~\cite{deng2022anomaly}, UniAD~\cite{you2022unified}, PatchCore~\cite{DBLP:conf/cvpr/RothPZSBG22}, CS-Flow~\cite{DBLP:conf/wacv/RudolphWRW22}, and STPM~\cite{salehi2021multiresolution}. SPADE, U-Std, PaDim, MKDAD, MAD, UniAD, CS-Flow, PatchCore, and STPM employ pre-trained models. U-Std, MKDAD, and STPM introduce knowledge distillation. DAAD+ and RIAD are image reconstruction methods. DRAEM, SGSF, and Cutpaste introduce human prior information to forge anomalous samples participating in the training phase. MF and UniAD are based on the transformer \cite{DBLP:conf/iclr/DosovitskiyB0WZ21} architecture.

 \begin{table*}[]
\scriptsize
	\centering
	\renewcommand\arraystretch{1.2}
	\caption{
		The anomaly detection results in terms of $\mathcal{AUROC}_{det}$ (\%) on MVTec-AD~\cite{bergmann2019mvtec}. The best results are marked in bold. An average score over all classes is reported in the last column (\textbf{MEAN}).}
	\resizebox{\textwidth}{!}{%
		\setlength{\tabcolsep}{1mm}{
			\begin{tabular}{r|ccccc|c|cccccccccc|c|c}
				\hline
				\multirow{2}{*}{Methods} & \multicolumn{5}{c}{Textures}                                                &                        & \multicolumn{10}{c}{Objects}                                                                                                                               &                        & \multirow{2}{*}{\textbf{MEAN}} \\ \cline{2-18}
				& Carpet        & Grid         & Leather      & Tile          & Wood          & \textit{\textbf{mean}} & Bottle       & Cable         & Capsule       & Hazelnut      & Metal nut     & Pill          & Screw         & Toothbrush   & Transistor    & Zipper       & \textit{\textbf{mean}} &                                \\ \cline{1-1} \cline{2-19} 
				
				SPADE      \cite{cohen2020sub}              & -             & -            & -            & -             & -             & -                      & -            & -             & -             & -             & -             & -             & -             & -            & -             & -            & -                      & 85.5                           \\
				U-Std   \cite{bergmann2020uninformed}                 & 91.6          & 81.0         & 88.2         & 99.1          & 97.7          & 91.5                   & 99.0         & 86.2          & 86.1          & 93.1          & 82            & 87.9          & 54.9          & 95.3         & 81.8          & 91.9         & 85.8                   & 87.7                           \\
				MKDAD     \cite{salehi2021multiresolution}               & 79.3          & 78           & 95.1         & 91.6          & 94.3          & 87.7                   & 99.4         & 89.2          & 80.5          & 98.4          & 73.6          & 82.7          & 83.3          & 92.2         & 85.6          & 93.2         & 87.8                   & 87.7                           \\
				DAAD+     \cite{hou2021divide}               & 86.6          & 95.7         & 86.2         & 88.2          & 98.2          & 91.0                   & 97.6         & 84.4          & 76.7          & 92.1          & 75.8          & 90            & 98.7          & 99.2         & 87.6          & 85.9         & 88.8                   & 89.5                           \\
				RIAD     \cite{zavrtanik2021reconstruction}               & 84.2          & 99.6         & \textbf{100.0} & 98.7          & 93            & 95.1                   & 99.9         & 81.9          & 88.4          & 83.3          & 88.5          & 83.8          & 84.5          & \textbf{100.0} & 90.9          & 98.1         & 89.9                   & 91.7                           \\
				MAD    \cite{rippel2021modeling}                  & 95.5          & 92.9         & \textbf{100.0} & 97.4          & 97.6          & 96.7                   & \textbf{100.0} & 94.0          & 92.3          & 98.7          & 93.1          & 83.4          & 81.2          & 95.8         & 95.9          & 97.9         & 93.2                   & 94.4                           \\
				Cutpaste     \cite{li2021cutpaste}            & 93.1          & 99.9         & \textbf{100.0} & 93.4          & 98.6          & 97.0                   & 98.3         & 80.6          & 96.2 & 97.3          & 99.3          & 92.4          & 86.3          & 98.3         & 95.5          & 99.4         & 94.4                   & 95.2                           \\
				MF \cite{wu2021learning}&94.0 &85.9&99.2&99.0&\textbf{99.2}&95.5&99.1&97.1&87.5&99.4&96.2&90.1&97.5&\textbf{100.0}&94.4&98.6&96.0&95.8\\
				PaDim      \cite{defard2021padim}              & 99.8 & 96.7         & \textbf{100.0} & 98.1          & \textbf{99.2} & 98.8                   & 99.9         & 92.7          & 91.3          & 92.0          & 98.7          & 93.3          & 85.8          & 96.1         & 97.4          & 90.3         & 93.8                   & 95.5                           \\
				%AnoSeg      \cite{DBLP:journals/corr/abs-2110-03396}             & 96.0          & 99.0         & 99.0         & 98.0          & 99.0          & 98.2                   & 98.0         & \textbf{98.0} & 84.0          & 98.0          & 95.0          & 87.0          & 97.0 & 99.0         & 96.0          & \textbf{99.0}         & 95.1                   & 96.0                           \\ 
				UniAD \cite{you2022unified} &99.9&98.5&\textbf{100.0}&99.0&97.9&99.0&\textbf{100.0}&97.6&85.3&99.9&99.0&88.3&91.9&95.0&\textbf{100.0}&96.7&95.3& 96.6 \\ 
                CS-Flow~\cite{DBLP:conf/wacv/RudolphWRW22} &99.3&99.0&99.7&98.0&96.7&98.5 &99.0&97.1&98.6&98.9&98.2&98.5&98.9&98.9&80.5&99.1&96.8&97.4\\
                
                PatchCore~\cite{DBLP:conf/cvpr/RothPZSBG22}  & -             & -            & -            & -             & -             & -                      & -            & -             & -             & -             & -             & -             & -             & -            & -             & -            & -                      & \textbf{99.0}\\ \hdashline
                
				\textit{\textbf{ReDi}}            & \textbf{100.0}          & \textbf{100.0} & \textbf{100.0} & \textbf{99.6} & 98.9          & \textbf{99.7}          & \textbf{100.0} & 97.9          & \textbf{96.4}          & \textbf{100.0} & \textbf{100.0} & \textbf{97.6} &\textbf{98.3}          & 98.6         & 97.0 & 97.8& \textbf{98.4}          & 98.8  \\ \hline
	\end{tabular}}}
	\label{TAB.1} 
\end{table*}

\begin{table*}[]
\scriptsize
	\centering
	\renewcommand\arraystretch{1.4}
	\caption{
		The anomaly segmentation results in terms of "{\color[HTML]{34696D} \tiny $\mathcal{AUROC}_{seg}$} / $\mathcal{AP}_{seg}$" (\%) on MVTec-AD~\cite{bergmann2019mvtec}.}
	\resizebox{\textwidth}{!}{%
		\setlength{\tabcolsep}{1mm}{
			\resizebox{\columnwidth}{!}{%
		\begin{tabular}{c|c|ccccccccccc}
			\hline
			&            & SPADE   \cite{cohen2020sub}                                            & U-std   \cite{bergmann2020uninformed}                          & RIAD \cite{zavrtanik2021reconstruction}                             & Padim    \cite{defard2021padim}                        & UniAD      \cite{you2022unified}                     & STPM   \cite{salehi2021multiresolution} &
            PatchCore~\cite{DBLP:conf/cvpr/RothPZSBG22}
   & RD    \cite{deng2022anomaly}        &RD++~\cite{Tien_2023_CVPR}     &RN~\cite{Gu_2023_ICCV}             & \textit{\textbf{ReDi}}                             \\ \hline
   From&&Arxiv&CVPR2020 &PR2021&ICPR2021&NeurIPS2022&CVPR2021&CVPR2022&CVPR2022&CVPR2023&ICCV2023&-\\ \hline
			& Carpet     & {\color[HTML]{34696D} \tiny 97.5} / - & {\color[HTML]{34696D} \tiny 93.5} / 52.2 & {\color[HTML]{34696D}\tiny 96.3} /  61.4 & {\color[HTML]{34696D} \tiny 99.0} /  60.7 & {\color[HTML]{34696D}\tiny 98.0} /  - & {\color[HTML]{34696D}\tiny 99.1} /  65.3 & {\color[HTML]{34696D}\tiny-} / - & {\color[HTML]{34696D}\tiny 98.9} /  64.1 & {\color[HTML]{34696D}\tiny99.2} / 64.3 & {\color[HTML]{34696D}\tiny-} / -& {\color[HTML]{34696D}\tiny 99.2} /  \textbf{68.4} \\
			& Grid       & {\color[HTML]{34696D}\tiny 93.7} /  -   & {\color[HTML]{34696D} \tiny 89.9} /  10.1 & {\color[HTML]{34696D}\tiny 98.8} /  36.4 & {\color[HTML]{34696D}\tiny 98.6} /  35.7 & {\color[HTML]{34696D} 94.6} /  - & {\color[HTML]{34696D}\tiny 99.1} /  45.4 & {\color[HTML]{34696D}\tiny-} / -& {\color[HTML]{34696D}\tiny 99.3} /  47.6 & {\color[HTML]{34696D}\tiny99.3} / 50.1& {\color[HTML]{34696D}\tiny-} / -& {\color[HTML]{34696D} \tiny 99.3} /  \textbf{50.6} \\
			& Leather & {\color[HTML]{34696D}\tiny 97.6} /  -  & {\color[HTML]{34696D}\tiny 97.8} /  40.9 & {\color[HTML]{34696D}\tiny 99.4} /  49.1 & {\color[HTML]{34696D}\tiny 99.0} /  \textbf{53.5} & {\color[HTML]{34696D}\tiny 98.3} /  - & {\color[HTML]{34696D}\tiny 99.2} /  42.9 & {\color[HTML]{34696D}\tiny-} / -& {\color[HTML]{34696D}\tiny 99.4} /  52.4 & {\color[HTML]{34696D}\tiny99.4} / 51.3 & {\color[HTML]{34696D}\tiny-} / -& {\color[HTML]{34696D}\tiny 99.5} /  52.3 \\
			& Tile       & {\color[HTML]{34696D}\tiny 88.5} / -   & {\color[HTML]{34696D}\tiny 92.1} / \textbf{53.3} & {\color[HTML]{34696D}\tiny 85.8} / 38.2 & {\color[HTML]{34696D}\tiny 94.1} / 46.3 & {\color[HTML]{34696D} \tiny 91.8} / - & {\color[HTML]{34696D}\tiny 95.2} / 47.0 & {\color[HTML]{34696D}\tiny-} / -& {\color[HTML]{34696D}\tiny 95.6} / 50.4 & {\color[HTML]{34696D}\tiny96.2} / 54.4 & {\color[HTML]{34696D}\tiny-} / -& {\color[HTML]{34696D}\tiny 95.7} / 49.5 \\
			& Wood & {\color[HTML]{34696D} \tiny 87.4} / -  & {\color[HTML]{34696D} \tiny 92.5} / \textbf{65.3} & {\color[HTML]{34696D} \tiny 89.1} / 52.6 & {\color[HTML]{34696D} \tiny 94.1} / 52.4 & {\color[HTML]{34696D} \tiny 93.4} / - & {\color[HTML]{34696D} \tiny 96.6} / 61.7 & {\color[HTML]{34696D}\tiny-} / -& {\color[HTML]{34696D} \tiny 95.3} / 53.6 & {\color[HTML]{34696D}\tiny95.6} / 52.6& {\color[HTML]{34696D}\tiny-} / -& {\color[HTML]{34696D} \tiny 98.7} / 55.0 \\ \cline{2-13}
			\multirow{-8}{*}{\rotatebox{270}{Texture}} & \textbf{mean}       & {\color[HTML]{34696D}\tiny 92.9} / -   & {\color[HTML]{34696D}\tiny 93.2} / 44.4 & {\color[HTML]{34696D}\tiny 93.9} / 47.5 & {\color[HTML]{34696D}\tiny 97.0} / 49.7 & {\color[HTML]{34696D}\tiny 95.2} / - & {\color[HTML]{34696D}\tiny 97.8} / 52.5& {\color[HTML]{34696D}\tiny-} / - & {\color[HTML]{34696D}\tiny 97.7} / 53.6 & {\color[HTML]{34696D}\tiny97.9} / 54.5& {\color[HTML]{34696D}\tiny-} / -& {\color[HTML]{34696D}\tiny 97.8} / \textbf{55.2} \\ \hdashline
			& Bottle     & {\color[HTML]{34696D}\tiny 98.4} / -   & {\color[HTML]{34696D}\tiny 97.8} / 74.2 & {\color[HTML]{34696D}\tiny 98.4} / 76.4 & {\color[HTML]{34696D}\tiny 98.2} / 77.3 & {\color[HTML]{34696D}\tiny 98.1} / - & {\color[HTML]{34696D}\tiny 98.8} / 80.6 & {\color[HTML]{34696D}\tiny-} / -& {\color[HTML]{34696D}\tiny 98.7} / 79.4 & {\color[HTML]{34696D}\tiny98.7} / 79.7& {\color[HTML]{34696D}\tiny-} / -& {\color[HTML]{34696D}\tiny 98.9} / \textbf{81.5} \\
			& Cable      & {\color[HTML]{34696D} \tiny 97.2} / -   & {\color[HTML]{34696D} \tiny 91.9} / 48.2 & {\color[HTML]{34696D} \tiny 84.2} / 24.4 & {\color[HTML]{34696D} \tiny 96.7} / 45.4 & {\color[HTML]{34696D} \tiny 96.8} / - & {\color[HTML]{34696D} \tiny 94.8} / 58.0 & {\color[HTML]{34696D}\tiny-} / -& {\color[HTML]{34696D} \tiny 97.4} / 59.2 & {\color[HTML]{34696D}\tiny98.3} / 61.7& {\color[HTML]{34696D}\tiny-} / -& {\color[HTML]{34696D} \tiny 97.9} / \textbf{72.6} 
			 \\
			& Capsule    & {\color[HTML]{34696D}\tiny 99.0} / -  & {\color[HTML]{34696D}\tiny 96.8} / 25.9 & {\color[HTML]{34696D}\tiny 92.8} / 38.2 & {\color[HTML]{34696D}\tiny 98.6} / \textbf{46.7} & {\color[HTML]{34696D}\tiny 97.9} / - & {\color[HTML]{34696D}\tiny 98.2} / 35.9 & {\color[HTML]{34696D}\tiny-} / -& {\color[HTML]{34696D}\tiny 98.7} / 45.8 & {\color[HTML]{34696D}\tiny98.7} / 47.1 & {\color[HTML]{34696D}\tiny-} / -& {\color[HTML]{34696D}\tiny 98.7} / 42.7 \\
			& Hazelnut   & {\color[HTML]{34696D} \tiny 99.1} / -  & {\color[HTML]{34696D} \tiny 98.2} / 57.8 & {\color[HTML]{34696D} \tiny 96.1} / 33.8 & {\color[HTML]{34696D} \tiny 98.1} / 61.1 & {\color[HTML]{34696D} \tiny 98.8} / - & {\color[HTML]{34696D} \tiny 98.9} / 60.3 & {\color[HTML]{34696D}\tiny-} / -& {\color[HTML]{34696D} \tiny 98.9} / 64.5 & {\color[HTML]{34696D}\tiny99.1} / 65.7& {\color[HTML]{34696D}\tiny-} / -& {\color[HTML]{34696D} \tiny 99.3} / \textbf{76.4} 
			 \\
			& Metal\_nut & {\color[HTML]{34696D}\tiny 98.1} / -  & {\color[HTML]{34696D}\tiny 97.2} / 83.5 & {\color[HTML]{34696D}\tiny 92.5} / 64.3 & {\color[HTML]{34696D}\tiny 97.3} / 77.4 & {\color[HTML]{34696D}\tiny 95.7} / - & {\color[HTML]{34696D}\tiny 97.2} / 79.3 & {\color[HTML]{34696D}\tiny-} / -& {\color[HTML]{34696D}\tiny 97.3} / 80.9 & {\color[HTML]{34696D}\tiny97.9} / 83.5& {\color[HTML]{34696D}\tiny-} / & {\color[HTML]{34696D}\tiny 98.0} / \textbf{88.9} 
			 \\
		& Pill       & {\color[HTML]{34696D}\tiny 96.5} / -  & {\color[HTML]{34696D} \tiny 96.5} / 62.0 & {\color[HTML]{34696D} \tiny 95.7} / 51.6 & {\color[HTML]{34696D} \tiny 95.7} / 61.2 & {\color[HTML]{34696D} \tiny 95.1} / - & {\color[HTML]{34696D} \tiny 94.7} / 63.3 & {\color[HTML]{34696D}\tiny-} / -& {\color[HTML]{34696D} \tiny 98.2} / \textbf{80.0} & {\color[HTML]{34696D}\tiny98.4} / 79.8& {\color[HTML]{34696D}\tiny-} / -& {\color[HTML]{34696D} \tiny 98.4} / 79.4 
		\\
			& Screw  & {\color[HTML]{34696D} \tiny 98.9} / -  & {\color[HTML]{34696D} \tiny 97.4} / 7.8 & {\color[HTML]{34696D} \tiny 98.8} / 43.9 & {\color[HTML]{34696D} \tiny 94.4} / 21.7 & {\color[HTML]{34696D} \tiny 97.4} / - & {\color[HTML]{34696D} \tiny 98.6} / 26.9 & {\color[HTML]{34696D}\tiny-} / -& {\color[HTML]{34696D} \tiny 99.3} / \textbf{54.8} & {\color[HTML]{34696D}\tiny99.6} / 55.6& {\color[HTML]{34696D}\tiny-} / -& {\color[HTML]{34696D} \tiny 99.6} / 44.8 \\
			   & Toothbrush  & {\color[HTML]{34696D} \tiny 97.9} / -  & {\color[HTML]{34696D} \tiny 97.9} / 37.7 & {\color[HTML]{34696D} \tiny 98.9} / 50.6 & {\color[HTML]{34696D} \tiny 98.8} / 54.7 & {\color[HTML]{34696D} \tiny 97.8} / - & {\color[HTML]{34696D} \tiny 98.9} / 48.8 & {\color[HTML]{34696D}\tiny-} / -& {\color[HTML]{34696D} \tiny 99.1} / 54.5 & {\color[HTML]{34696D}\tiny99.1} / 56.2& {\color[HTML]{34696D}\tiny-} / -& {\color[HTML]{34696D} \tiny 98.9} / \textbf{62.4}
				 \\
			& Transistor & {\color[HTML]{34696D} \tiny 94.1} / -  & {\color[HTML]{34696D} \tiny 73.7} / 27.1 & {\color[HTML]{34696D} \tiny 87.7} / 39.2 & {\color[HTML]{34696D} \tiny 97.6} / \textbf{72.0} & {\color[HTML]{34696D} \tiny 98.7} / - & {\color[HTML]{34696D} \tiny 81.9} / 44.4 & {\color[HTML]{34696D}\tiny-} / -& {\color[HTML]{34696D} \tiny 92.5} / 55.7 & {\color[HTML]{34696D}\tiny94.8} / 59.1& {\color[HTML]{34696D}\tiny-} / -& {\color[HTML]{34696D} \tiny 96.1} / 70.1 \\
			& Zipper & {\color[HTML]{34696D} \tiny 96.5} / -  & {\color[HTML]{34696D} \tiny 95.6} / 36.1 & {\color[HTML]{34696D} \tiny 97.8} / \textbf{63.4} & {\color[HTML]{34696D} \tiny 98.4} / 58.2 & {\color[HTML]{34696D} \tiny 96.0} / - & {\color[HTML]{34696D} \tiny 98.0} / 54.9 & {\color[HTML]{34696D}\tiny-} / -& {\color[HTML]{34696D} \tiny 98.2} / 60.6 & {\color[HTML]{34696D}\tiny98.8} / 61.1& {\color[HTML]{34696D}\tiny-} / -& {\color[HTML]{34696D} \tiny 98.9} / 53.3  \\ \cline{2-13}
			\multirow{-12}{*}{\rotatebox{270}{Object}} & \textbf{mean} & {\color[HTML]{34696D} \tiny 97.6} / -  & {\color[HTML]{34696D} \tiny 94.3} / 46.0 & {\color[HTML]{34696D} \tiny 94.3} / 48.6 & {\color[HTML]{34696D} \tiny 97.4} / 57.6 & {\color[HTML]{34696D} \tiny 97.3} / - & {\color[HTML]{34696D} \tiny 96.0} / 55.2 & {\color[HTML]{34696D}\tiny-} / -& {\color[HTML]{34696D} \tiny 97.8} / 63.5 & {\color[HTML]{34696D}\tiny98.3} / 65.0& {\color[HTML]{34696D}\tiny-} / -& {\color[HTML]{34696D} \tiny 98.5} / \textbf{67.2} \\ \hdashline
			\multicolumn{2}{c}{\textbf{Mean}} & {\color[HTML]{34696D} \tiny 96.5} / - & {\color[HTML]{34696D} \tiny 93.9} / 45.5 & {\color[HTML]{34696D} \tiny 94.2} / 48.2 & {\color[HTML]{34696D} \tiny 97.4} / 55.0 & {\color[HTML]{34696D} \tiny 96.6} / - & {\color[HTML]{34696D} \tiny 96.6} / 54.3 & {\color[HTML]{34696D}\tiny 98.1} / 56.1& {\color[HTML]{34696D} \tiny 97.8} / 60.2& {\color[HTML]{34696D}\tiny98.2} / 61.5& {\color[HTML]{34696D}\tiny98.1} / 59.6 & {\color[HTML]{34696D} \tiny 98.3} / \textbf{63.2} \\
			 \hline
		\end{tabular}%
	}}}\label{TAB.2} 
\end{table*}

%------------------------------------------------------------------------
\subsection{Comparisons with Existing Methods}

\begin{figure*}[t]
	\centering
	\includegraphics[width=0.99\linewidth]{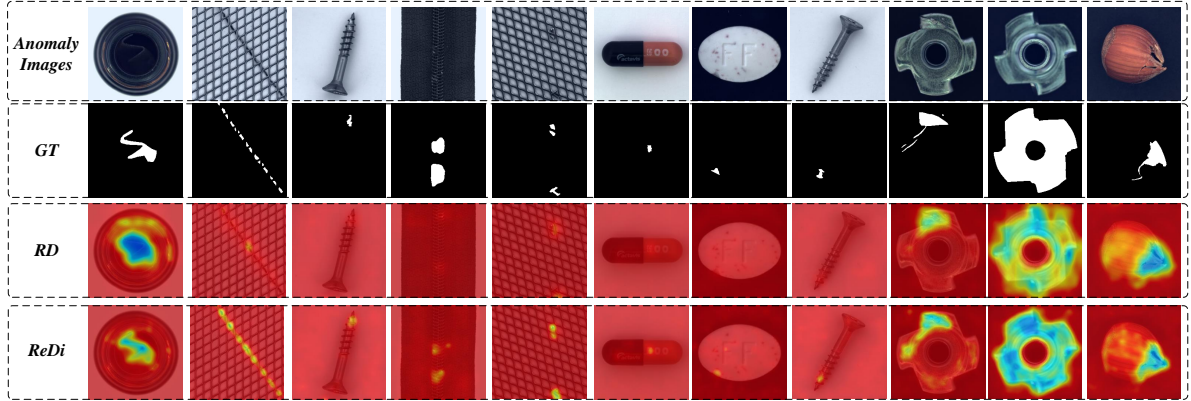}
	\caption{Comparison of the anomaly segmentation results of the proposed \textit{ReDi} with  RD. \textit{ReDi} has a more powerful anomaly segmentation ability, effectively segmenting tiny anomaly regions.}
	\label{fig:vis5} \vspace{-0.3cm}
\end{figure*}

Quantitative results of anomaly detection on MVTec-AD are shown in Table~\ref{TAB.1}. Among the 15 classes on MVTec-AD, the proposed \textit{ReDi} achieves superior performance on 10 classes. It is close to current state-of-the-art (SOTA) methods such as PatchCore, but \textit{ReDi} has a significantly stronger anomaly segmentation ability. For Texture and Object, \textit{ReDi} establishes a new state-of-the-art performance with 99.7\% and 98.4\%, respectively. Even when compared with the current state-of-the-art transformer-based methods, such as UniAD and MF, \textit{ReDi} obtains the most advanced anomaly detection performance by first recovering and then discriminating. Compared to existing anomaly detection methods with self-supervised training, our ReDi recovers the original image through the corresponding self-generated map, which is an efficient method that can guarantee a large anomaly recovery error while keeping the normal region small, avoiding the ``equivalent shortcut'' problem. The overall result demonstrates the effectiveness of \textit{ReDi} in introducing low-level information and normal semantic information to recover images and discriminate anomalies in the feature space.

\begin{table*}[h]
\scriptsize
% \color{red}
\centering
\caption{{Comparison of the proposed method with recent methods based on diffusion model~\cite{ho2020denoising,song2020denoising} and CLIP~\cite{radford2021learning}.}}
\begin{tabular}{c|cccccc}
\hline
&AnoCLIP~\cite{zhou2023anomalyclip} & DiffusionAD~\cite{zhang2023diffusionad} & Destseg~\cite{zhang2023destseg} & RealNet~\cite{zhang2024realnet} & \textit{ReDi} \\ \hline
$\mathcal{AUROC}_{det}$&
91.5           & 99.7               & 98.6           & 99.6           & 98.8 \\
$\mathcal{AUROC}_{seg}$&91.1           & 98.7               & 97.9           & 99.0           & 98.3 \\
$\mathcal{PRO}_{seg}$&81.4           & 95.7               & -              & -              & 95.2 \\ \hline
\end{tabular}
\label{tabsup3}
\end{table*}

\begin{wrapfigure}{r}{0.5\textwidth}
% \centering
\includegraphics[width=1\linewidth]{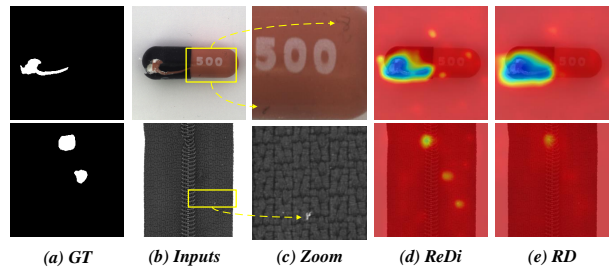}
	\caption{Samples with labeling inaccuracies. ``Zoom" denotes the zoomed-in view of the yellow boxes. Our \textit{ReDi} accurately segments unpredictable and subtle anomalous objects. }
	\label{fig:vis6}
 % \vspace{0.3cm}
\end{wrapfigure}

Quantitative results ($\mathcal{AUROC}{seg}$ and $\mathcal{AP}{seg}$) on MVTec-AD are shown in Table \ref{TAB.2}. With Table \ref{TAB.1}, it can be found that the current advanced methods can solve the anomaly detection problem well, but are not effective for anomaly segmentation. \textit{ReDi} achieves a new SOTA with 98.3\% and 63.2\% in both $\mathcal{AUROC}{seg}$ and $\mathcal{AP}{seg}$, respectively. {In particular, it outperforms the first runner-up method RD by 3.7\% and the third runner-up method PaDim by 10\% in terms of $\mathcal{AP}{seg}$ on anomalous segmentation for Object, which is difficult to solve by these methods.} Although Patchcore is effective in solving anomaly detection, ReDi outperforms it in anomaly segmentation by 7\% in terms of $\mathcal{AP}{seg}$. The \textit{ReDi} method may solve both high-confidence anomaly detection and anomaly segmentation. Figure \ref{fig:vis5} shows the visualization results of anomalous segmentation of anomaly images. It can be found that fine anomalies in the ``Screw" and ``Grid" classes can be segmented, while the boundary of the anomaly segmentation is more reasonable. For example, drugs and nails, some minor abnormalities are not segmented in RD. One possible reason is that the student model of RD can extract the same features as the teacher model, resulting in the ``identity shortcut'', while \textit{ReDi} utilizes HOG images to explicitly introduce low-level information prompts to ensure the details of recovery and then segment the difference regions(\ie, anomaly regions). Besides, as shown in Figure \ref{fig:vis6}, we find an interesting phenomenon that can explain the reduced performance of some classes (\eg, ``Capsule"). There are some unlabeled ``anomalous objects'' in anomaly images, \textit{ReDi} can detect these anomalous regions. Since \textit{ReDi} takes the normal image as the prompt image, the normal semantic features guide the recovery results without ``abnormal objects". Therefore, \textit{ReDi} can detect anomalies that differ from normal patterns. 

% \begin{figure}
% 	\centering
% 	\includegraphics[width=0.7\linewidth]{picture/Figure5}
% 	\caption{Recovery results for four different self-generated maps in Recover Network. The top line is the Input/recovery images, and the bottom line is the GT/self-generated maps.}
% 	\label{fig:figure6}
% \end{figure}

% \begin{table}[]
% \tiny
% \centering
% \renewcommand\arraystretch{1.2}
% 	\caption{Result comparisons of anomaly detection and anomaly segmentation of different self-generated maps on MVTec-AD~\cite{bergmann2019mvtec}. }{
		
% 			\setlength{\tabcolsep}{0.6mm}{
% \begin{tabular}{@{}p{2.0cm}<{\centering}|p{1.6cm}<{\centering}p{3.5cm}<{\centering}@{}}
% \toprule
% Features &$\mathcal{AUROC}_{det}$ & $\mathcal{AUROC}_{seg}$ / $\mathcal{AP}_{seg}$  \\ \hline
% Canny \cite{Canny86a} &93.4 &97.7 / 58.8\\
% Sobel \cite{duda1973pattern}   & 97.7  & 98.0 / 62.6 \\
% Roberts \cite{roberts1963machine}   & 97.9  & 98.0 / 62.1 \\
% HOG  \cite{dalal2005histograms}    & 98.8  & 98.2 / 63.2 \\ \hline
% \end{tabular}%
% }} \label{tab:tab5}
% \end{table}

% \begin{figure}[htbp]
% \centering
% \begin{minipage}[t]{0.46\textwidth}

% \end{minipage}
% \begin{minipage}[t]{0.48\textwidth}

% \end{minipage}
% \end{figure}

Quantitative results of anomaly detection ($\mathcal{AUROC}_{det}$) and segmentation ($\mathcal{AUROC}_{seg}$ and $\mathcal{AP}_{seg}$) on  KolektorSDD2 dataset are shown in Table \ref{TAB.3}. \textit{ReDi} significantly outperforms recent AD methods in three metrics and achieves the new SOTA. \textit{ReDi} fully exploits the relationship of normal samples and helps Recover Network to filter unlearned anomalies by normal sample prompt. It demonstrates the effectiveness of using normal samples as prompt information for recovery and discriminating anomalous differences in the feature space.

\textcolor{black}{
In addition, we add a comparison of the proposed \textit{ReDi} with recent methods introducing advanced models~(such as the Diffusion models~\cite{ho2020denoising,song2020denoising} and CLIP~\cite{radford2021learning}). Meanwhile, we introduce the per-region-overlap (PRO) metric (denoted as $\mathcal{PRO}_{seg}$)~\cite{zhou2023anomalyclip,zhang2023diffusionad}. Unlike $\mathcal{AUROC}_{seg}$, which is used for pixel-by-pixel measurements, the PRO score treats anomalous regions of any size equally. The advanced methods compared are as follows:
AnoCLIP~\cite{zhou2023anomalyclip}: it introduces a powerful CLIP text encoder and image encoder~\cite{radford2021learning}.
DiffusionAD~\cite{zhang2023diffusionad}: it introduces an advanced diffusion model~\cite{ho2020denoising,song2020denoising} for anomaly detection.
Destseg~\cite{zhang2023destseg}: it uses prior knowledge to additionally construct forged anomaly samples to train supervised segmented anomaly networks.
RealNet~\cite{zhang2024realnet}: it introduces the diffusion model~\cite{ho2020denoising,song2020denoising} to generate forged anomaly samples for anomaly detection.
Among these methods, they all use state-of-the-art models, especially the recent advanced diffusion model structure. A large model leads to a significant increase in the number of parameters, computation, and inference time. We report in Table~\ref{tabsup3} the results of \textit{ReDi} compared to these methods on the MVTec AD dataset. The experimental results show that our method remains competitive in anomaly detection and anomaly localization tasks. In the future, we will explore the use of advanced diffusion model structures to improve the HIP of \textit{ReDi}.}

%The visualization of anomaly segmentation, comparing \textit{ReDi} with  RD, is shown in Figure \ref{fig:vis5}. First, the heat map results show that \textit{ReDi} has a powerful anomaly segmentation capability. Second, \textit{ReDi} is more powerful in capturing tiny anomaly regions. Third, \textit{ReDi} is able to ensure a lower false positive rate for normal regions. In addition, we find an interesting experimental phenomenon that can explain the reduced performance of some classes. Anomalous products are industrial products without defects, thus the pixel annotations are all in the product region. However, there are some details in these high-resolution images where background anomalies or anomalies of unpredictable objects are unlabeled, as shown in Figure \ref{fig:vis6}. \textit{ReDi} can effectively detect unlabeled anomalies, which is advantageous, however $\mathcal{AUROC}_{seg}$ and $\mathcal{AP}_{seg}$ show reduced performance. The heatmaps in Figure \ref{fig:vis6} show that \textit{ReDi} focuses on all the anomalous patterns that are different from the normal samples and fit the requirements of the anomaly detection task.

\begin{table}[t!]
\scriptsize
\parbox{.49\linewidth}{
\centering
\caption{Anomaly detection and anomaly segmentation results on KolektorSDD2~\cite{bovzivc2021mixed}. The proposed \textit{ReDi} approach achieves new SOTA performance without any forged anomalous samples. }{
		% \resizebox{\columnwidth}{!}{%
			\setlength{\tabcolsep}{0.8mm}{
				\begin{tabular}{r|c|c|p{1.3cm}<{\centering}}
					\hline Methods& $\mathcal{AUROC}_{det}$& $\mathcal{AUROC}_{seg}$& $\mathcal{AP}_{seg}$           \\ \hline
					%\begin{tabular}[c]{@{}c@{}}Semi-\\ orthogonal\end{tabular}\cite{kim2021semi}
					Semi-orthogonal \cite{kim2021semi} & -                              & 98.1                           & -                              \\
					PaDim \cite{defard2021padim}                                                                                                                           & -                              & 95.6                           & -                              \\
					U-Std \cite{bergmann2020uninformed}                                                                                                                            & -& 95.0 & - \\
					DRAEM \cite{zavrtanik2021draem}                                                                                                                           & 93.1                           & 93.4                           & 50.5                           \\
					SGSF     \cite{xing2022self}                                                                                                                          & 93.5                           & 91.5                           & 51.6                           \\
					RD\cite{deng2022anomaly}                                                                                                                                 & 94.8                           & 98.2                           & 47.7                           \\ \hline
					\textit{\textbf{ReDi}}                                                                           & \textbf{96.2} & \textbf{98.8} & \textbf{54.6} \\ \hline
				\end{tabular}
	}}
 % }
 \label{TAB.3} 
}
\hfill
\parbox{.49\linewidth}{
\centering
\caption{Result comparisons of anomaly detection and anomaly segmentation of different self-generated maps on MVTec-AD~\cite{bergmann2019mvtec}. }{
		
			\setlength{\tabcolsep}{0.6mm}{
\begin{tabular}{@{}p{2.0cm}<{\centering}|p{1.6cm}<{\centering}p{3.5cm}<{\centering}@{}}
\toprule
Features &$\mathcal{AUROC}_{det}$ & $\mathcal{AUROC}_{seg}$ / $\mathcal{AP}_{seg}$  \\ \hline
Canny \cite{Canny86a} &93.4 &97.7 / 58.8\\
Sobel \cite{duda1973pattern}   & 97.7  & 98.0 / 62.6 \\
Roberts \cite{roberts1963machine}   & 97.9  & 98.0 / 62.1 \\
HOG  \cite{dalal2005histograms}    & 98.8  & 98.2 / 63.2 \\ \hline
\end{tabular}%
}} \label{tab:tab5}
}
\end{table}

\begin{table*}[]
\scriptsize
	\centering
	\renewcommand\arraystretch{1.2}
	\caption{Experimental results of introducing self-correlation loss for different backbones on MVTec-AD~\cite{bergmann2019mvtec}.}{
	%	\resizebox{\columnwidth}{!}
  % {%
			\setlength{\tabcolsep}{5.5mm}{
		\begin{tabular}{c|cc|cc|cc}
			\hline
			& RD   & +$L_S$ & \begin{tabular}[c]{@{}c@{}}Wide-\\ResNet50\end{tabular} & +$L_S$ & \begin{tabular}[c]{@{}c@{}}\textit{Redi}\\ W/O. $L_S$\end{tabular} & +$L_S$ \\ \hline
			$\mathcal{AUROC}_{det}$ & 98.7 & 98.8  \textcolor{blue}{(+0.1)} & 94.3                                                   & 96.7  \textcolor{blue}{(+2.4)} & 97.3                                                    & 98.8  \textcolor{blue}{(+1.5)} \\
			$\mathcal{AUROC}_{seg}$ & 97.8 & 98.9 \textcolor{blue}{(+1.1)} & 94.6                                                   & 97.6 \tiny \textcolor{blue}{(+3.0)}& 97.6                                                    & 98.2  \textcolor{blue}{(+0.6)}\\
			$\mathcal{AP}_{seg}$   & 60.2 & 61.4 \textcolor{blue}{(+1.2)} & 54.3                                                   & 58.1  \textcolor{blue}{(+3.8)}& 61.5                                                    & 63.2  \textcolor{blue}{(+1.7)}\\ \hline
		\end{tabular} 
	\label{tab5-1}
	}}
\end{table*}
\subsection{Ablation Studies}
The purpose of our ablation studies is to investigate the experimental performance of our proposed method in the following two aspects: comparisons with different self-generated maps and the effects of the HOG and prompt image.

\begin{wrapfigure}{r}{0.5\textwidth}
\centering
\includegraphics[width=0.99\linewidth]{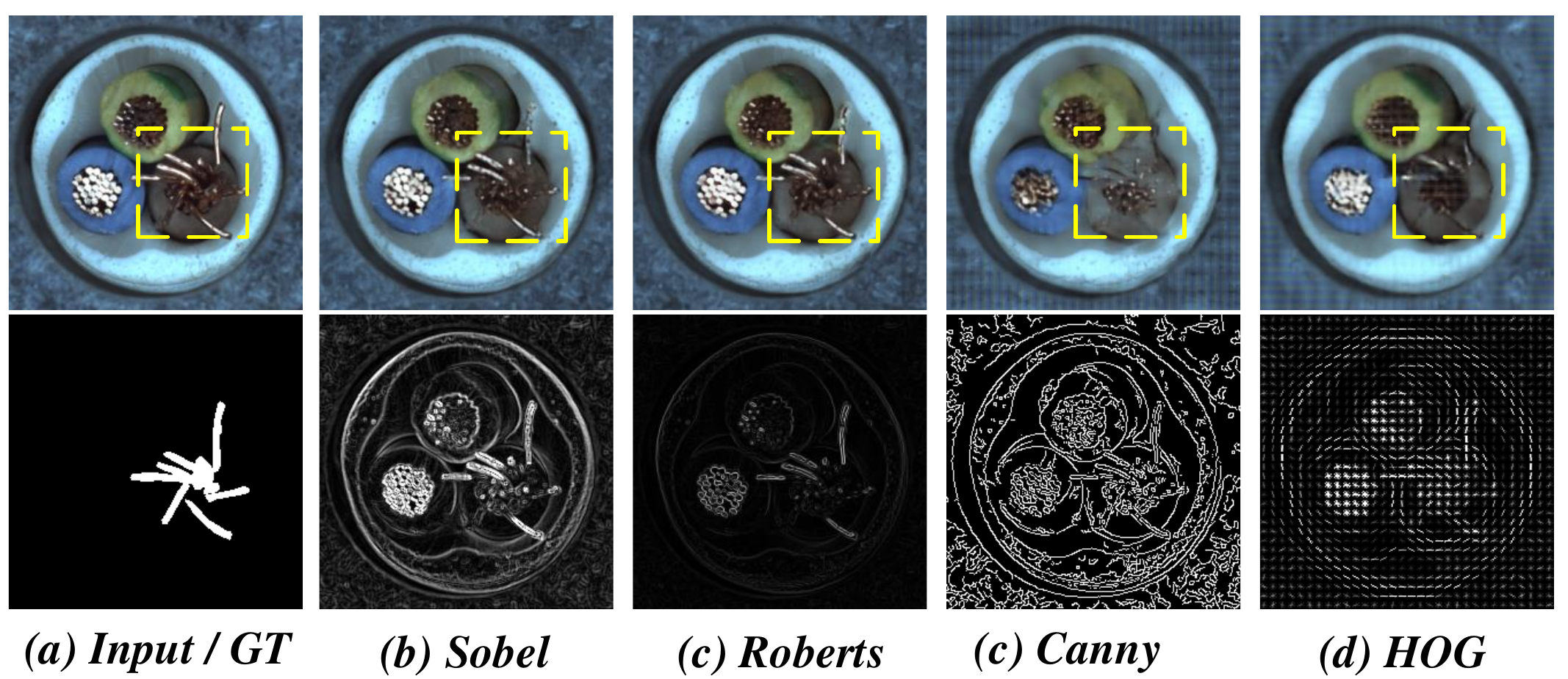}
	\caption{Recovery results for four different self-generated maps in Recover Network. The top is the Input/recovery images, and the bottom is the GT/self-generated maps.}
	\label{fig:figure6}
\end{wrapfigure}

\subsubsection{Comparisons with different self-generated maps} \label{431}

We explore the performance of four different self-generated feature maps, namely Sobel~\cite{duda1973pattern}, Canny~\cite{Canny86a}, Roberts~\cite{roberts1963machine}, and HOG~\cite{dalal2005histograms}, as inputs for the Recover Network. For the Sobel operator, we use \textit{kernelsize=3} using \textit{CV2.Sobel()}. For the Roberts operator, we use template [[-1, 0], [0, 1]] for x-axis and [[0, -1], [1, 0]] for y-axis. For the Canny operator, we use \textit{cv2.Canny(image, $th_1$, $th_2$)} to extract the canny edge. In this paper, we set $th_1=50$ and $th_2=100$. 
Figure \ref{fig:figure6} displays the recovery results when each of these four traditional features is used as self-generated maps. Within the region marked by the dashed box, we can observe that the edge features of Sobel and Roberts are able to effectively highlight the abnormal structural information, which consequently enables the accurate restoration of abnormal subjects. On the other hand, the Canny edge features struggle to maintain edge details, leading to substantial recovery errors in normal regions, even though the recovery error in abnormal regions is large. Of the four, the HOG map performs the best, a conclusion supported by the quantitative results in Table~\ref{tab:tab5}. These results suggest that some fine edge features might inadvertently lead to the generation of anomalous regions, providing an explanation as to why the model might easily learn the ``identical shortcut'' when the original image is used as an input. In other words, these features offer detailed cues, and the HOG, when used as a self-generated map, provides a better trade-off between the recovery results for normal and anomalous regions.

 %The performance of four different self-generated feature maps as input of Recover Network: Sobel~\cite{duda1973pattern}, Canny~\cite{Canny86a}, Roberts~\cite{roberts1963machine}, and HOG~\cite{dalal2005histograms} are explored. In Figure 5, we show the recovery results of each of the four traditional features as self-generated maps. It can be observed that in the region indicated by the dashed box, the Sobel and Roberts edge features are able to clearly label the abnormal structural information, which leads to the abnormal subjects can be accurately restored. canny edge features are difficult to ensure the edge details which leads to the large recovery error in the normal region, although the recovery error in the abnormal region is large. HOG as a self-generated map performs the best, which can be captured from the quantitative results in Table~\ref{tab:tab5}. This result is indicative that some fine edge features may lead to anomalous regions being generated, which may explain why the model can easily learn the ``idential shorcut'' when the original image (or the text above) is used as input, i.e., they provide a lot of detailed prompts, and HOG as self-generated map is better trade-off the recovery results for normal regions and anomalous regions.

\begin{table*}[]
\scriptsize
	\centering

	\renewcommand\arraystretch{1.7}
	\caption{The performance of \textit{ReDi} on MVTec-AD~\cite{bergmann2019mvtec} with different backbones of Recover Network. The result of HIP is optimal.}{
		\renewcommand\arraystretch{1.2}
		% \resizebox{\columnwidth}{!}
  \setlength{\tabcolsep}{4mm}
  {%
			\begin{tabular}{c|c|ccc|ccc}
				\hline
				& \multirow{2}{*}{Categories}          & \multicolumn{3}{c|}{	$\mathcal{AUROC}_{det}$} & \multicolumn{3}{c}{{\color[HTML]{34696D}\tiny	$\mathcal{AUROC}_{seg}$ }/ 	$\mathcal{AP}_{seg}$}                                                                                \\ \cline{3-8}
				&          & ImI  & IIHP    & HIP    & ImI                           & IIHP                                 & HIP                                  \\ \hline
				& Carpet     & 100.00      & 100.00  & 100.00 & {\color[HTML]{34696D}\tiny 99.20} / 67.85 & {\color[HTML]{34696D}\tiny 99.30} / 49.96 & {\color[HTML]{34696D}\tiny 99.20} / 68.42 \\
				& Grid       & 64.40       & 100.00  & 100.00  & {\color[HTML]{34696D}\tiny 84.80} / 12.55 & {\color[HTML]{34696D}\tiny 99.20} / 46.32 & {\color[HTML]{34696D}\tiny 99.30} / 50.63 \\
				& Leather    & 100.00      & 100.00  & 100.00 & {\color[HTML]{34696D}\tiny 99.40} / 52.56 & {\color[HTML]{34696D}\tiny 99.50} / 52.80 & {\color[HTML]{34696D}\tiny 99.50} / 52.30 \\
				& Tile       & 94.60       & 98.00   & 99.60 & {\color[HTML]{34696D}\tiny 92.90} / 45.53 & {\color[HTML]{34696D}\tiny 94.20} / 52.54 & {\color[HTML]{34696D}\tiny 94.70} / 49.95 \\
				
				& Wood       & 98.90       & 98.60   & 98.90  & {\color[HTML]{34696D}\tiny 94.10} / 50.48 & {\color[HTML]{34696D}\tiny 95.50} / 54.42 & {\color[HTML]{34696D}\tiny 95.70} / 54.99 \\  \cline{2-8}
				\multirow{-6}{*}{\rotatebox{270}{Texture}} & \textbf{mean}       & 91.58       & 99.32   & 99.70  & {\color[HTML]{34696D}\tiny 94.08} / 45.79 & {\color[HTML]{34696D}\tiny 97.54} / 51.21 & {\color[HTML]{34696D}\tiny 97.80} / 55.16 \\ \hdashline
				& Bottle     & 89.80       & 100.00  & 100.00  & {\color[HTML]{34696D}\tiny 93.80} / 51.94 & {\color[HTML]{34696D}\tiny 99.10} / 84.41 & {\color[HTML]{34696D}\tiny 98.90} / 81.54 \\
				& Cable      & 52.20       & 95.50   & 97.90  & {\color[HTML]{34696D}\tiny 71.40} / 10.36 & {\color[HTML]{34696D}\tiny 97.70} / 59.28 & {\color[HTML]{34696D}\tiny 97.90}  / 72.60\\
				& Capsule    & 59.40       & 89.90   & 96.40  & {\color[HTML]{34696D}\tiny 95.30} / 27.62 & {\color[HTML]{34696D}\tiny 97.40} / 41.17 & {\color[HTML]{34696D}\tiny 98.70} / 42.69 \\
				& Hazelnut   & 90.90       & 99.90   & 100.00  & {\color[HTML]{34696D}\tiny 98.40} / 61.06 & {\color[HTML]{34696D}\tiny 99.20} / 70.11 & {\color[HTML]{34696D}\tiny 99.30} / 76.43 \\
				& Metal\_nut & 90.30       & 99.30   & 100.00  & {\color[HTML]{34696D}\tiny 96.50} / 76.72 & {\color[HTML]{34696D}\tiny 97.60} / 82.72 & {\color[HTML]{34696D}\tiny 98.00} / 88.89 \\
				& Pill       & 73.30       & 94.50   & 97.60  & {\color[HTML]{34696D}\tiny 93.80} / 45.95 & {\color[HTML]{34696D}\tiny 97.40} / 73.10 & {\color[HTML]{34696D}\tiny 98.40} / 79.38 \\
				& Transistor & 92.00       & 98.80   & 97.00  & {\color[HTML]{34696D}\tiny 88.70} / 48.73 & {\color[HTML]{34696D}\tiny 93.70} / 57.75 & {\color[HTML]{34696D}\tiny 96.10} / 70.14 \\
				& Screw      & 48.90       & 90.20   & 98.30  & {\color[HTML]{34696D}\tiny 85.40} / 1.00  & {\color[HTML]{34696D}\tiny 98.10} / 42.66 & {\color[HTML]{34696D}\tiny 99.60} / 44.79 \\
				& Toothbrush & 91.10       & 91.40   & 98.60  & {\color[HTML]{34696D}\tiny 98.50} / 52.75 & {\color[HTML]{34696D}\tiny 98.60} / 59.16 & {\color[HTML]{34696D}\tiny 98.90} / 62.35 \\
				& Zipper     & 74.20       & 87.60   & 97.80  & {\color[HTML]{34696D}\tiny 89.10} / 20.19 & {\color[HTML]{34696D}\tiny 96.30} / 46.44 & {\color[HTML]{34696D}\tiny 98.90} / 53.32 \\ \cline{2-8}
				\multirow{-11}{*}{\rotatebox{270}{Object}} & \textbf{mean}       & 76.21       & 94.71   & 98.40  & {\color[HTML]{34696D}\tiny 91.09} / 39.53 & {\color[HTML]{34696D}\tiny 97.51} / 61.68 & {\color[HTML]{34696D}\tiny 98.50} / 67.21 \\ \hdashline
				\multicolumn{2}{c|}{\textbf{MEAN}}               & 81.33       & 96.25   & 98.81  & {\color[HTML]{34696D}\tiny 92.09} / 41.62 & {\color[HTML]{34696D}\tiny 97.52} / 58.18 & {\color[HTML]{34696D}\tiny 98.25} / 63.19 \\ \hline
			\end{tabular} \label{tab5-2}
	}}
\end{table*}

\subsubsection{The effects of the HOG and Prompt image}
To validate whether the HOG images are able to provide low-level information supporting the recovery of detailed information, we compare three different methods on MVTec-AD. Specifically, they are described as follows: 
\begin{itemize}
    \item \textbf{Image Inpainting (ImI).} 
    {A recovery model is trained using the image inpainting task to detect anomalies by pixel-level errors. During the training phase of ImI,  given an input image $X$ of size $H \times W$ and the corresponding HOG image $H$. We randomly eliminate a rectangular region of $X$ by a binary mask $M$, and $M_{ij}=0$ means that the information in $X$ with coordinate $(i,j)$ is eliminated. The input of IIHP is $X_H$.
\begin{equation}
	X_I=X \times M.
\end{equation}
    The optimization objective of ImI is to constrain the recovery of $X_I$ to $X$, as follows,
\begin{equation}
	\mathop {\min }\limits_{\theta_1}  {L_{Rec}}(Y = {f_{\theta_1} }(X_I),X),
\end{equation}
where $L_{Rec}$ is the recovery loss function and $f_{\theta_1}$ represents the parameters of the ImI model, which introduces the standard encoder-decoder structure of UNet.}

 %-------------------------------------------

\begin{figure}[]
\centering
\begin{minipage}[t]{0.49\textwidth}
\centering
\includegraphics[width=0.99\linewidth]{picture/figure_6}
	\caption{Comparison of recovery results of three methods. IIHP improves the recovery of details than ImI (red box), while the recovery of abnormal and normal in HIP is more as expected.}
	\label{fig:figure5}
\end{minipage}
\begin{minipage}[t]{0.1\textwidth}
\end{minipage}
\begin{minipage}[t]{0.40\textwidth}
\centering
\includegraphics[width=0.99\linewidth]{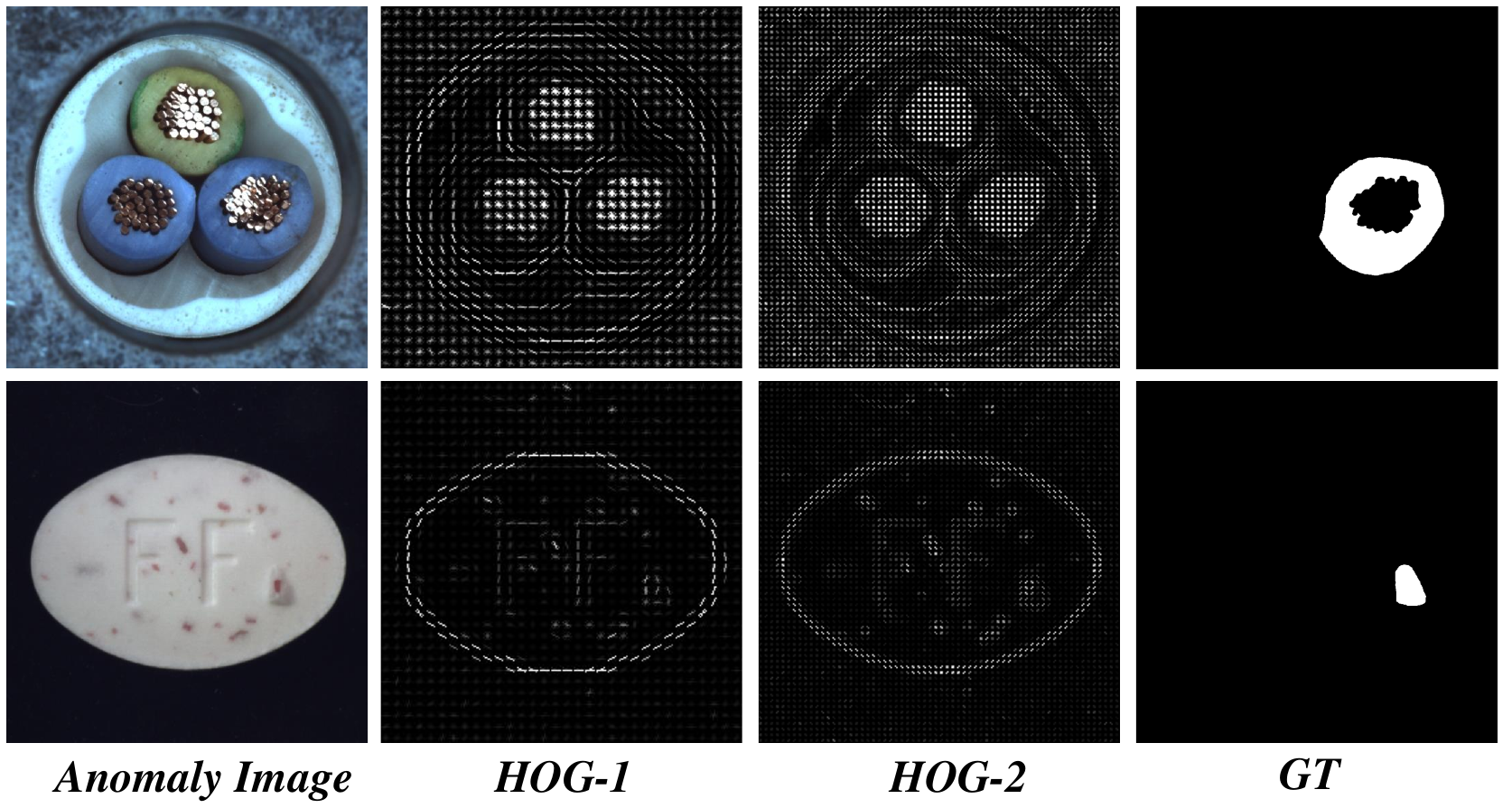}
		\caption{HOG images of the abnormal samples. ``HOG-1" indicates $bin=9$ and the local area $size$ is $8\times 8$. ``HOG-2" indicates $bin=18$ and the local area $size$ is $4\times 4$. It can be found that the ``HOG-2" prompt information is more detailed.}
		\label{fig:hog}
\end{minipage}
\end{figure}
 
 \item \textbf{Image Inpainting with HOG Prompt (IIHP)}.  {IIHP is an image inpainting-based method that introduces HOG images to provide low-level for the eliminated regions, which solves the problem of lack of details in image inpainting.  
During the training phase of IIHP,  given an input image $X$ of size $H \times W$ and the corresponding HOG image $H$. The input of IIHP is $X_H$.
\begin{equation}
	X_H=X \times M + {\rm{H}} \times (1 - M).
\end{equation}
The optimization objective of IIHP is to constrain the recovery of $X_H$ to $X$, as follows,
\begin{equation}
	\mathop {\min }\limits_{\theta_2}  {L_{Rec}}(Y = {f_{\theta_2} }(X_H),X),
\end{equation}
where $f_{\theta_2}$ represents the parameters of the IIHP model, which introduces the standard encoder-decoder structure of UNet.}
%---------------------------------------
\item \textbf{HOG with Image Prompt (HIP)}.{ The methodology is presented in in Section~\ref{sec:3.2}.}
\end{itemize}

% \begin{figure}
% 	\centering
% 	\includegraphics[width=0.99\linewidth]{picture/Figure6}
% 	\caption{Comparison of recovery results of three methods. IIHP improves the recovery of details than ImI (red box), while the recovery of abnormal and normal in HIP is more as expected.}
% 	\label{fig:figure5}
% \end{figure}

\begin{table}[]
\scriptsize
		\centering
		\renewcommand\arraystretch{1.4}
		\caption{The effect of the prompt image $X^+$ in HIP. ``Free'' indicates that no prompt images are utilized and the input contains only HOG images. ``Self-supervised'' and ``Pertrained'' represent the network trained by self-supervised reconstruction and the pre-trained network (WideResNet50) extracting features and generating $X^+$, respectively.}{
			\renewcommand\arraystretch{1.5}
			% \resizebox{\columnwidth}{!}{%
				\begin{tabular}{cc|lll}
					\hline
					Prompt Type     & Metric      & Texture            &    Object  	            &         MEAN        \\ \hline
					\multirow{3}{*}{Free}   & $\mathcal{AUROC}_{det}$ & 96.56          & 94.80          & 95.39          \\
					& $\mathcal{AUROC}_{seg}$   &  \textbf{97.86 }         & 91.78          & 93.81          \\
					& $\mathcal{AP}_{seg}$    & 48.81          & 59.06          & 55.64          \\ \hline
					\multirow{3}{*}{Self supervised}   & $\mathcal{AUROC}_{det}$ & 97.30  \textcolor{blue}{(+0.74)}        & 97.79  \textcolor{blue}{(+2.09)}        & 97.63    \textcolor{blue}{(+2.24)}      \\
					&$\mathcal{AUROC}_{seg}$   & 99.26  \textcolor{blue}{(+1.40)}        & 95.90    \textcolor{blue}{(+4.12)}      & 97.02  \textcolor{blue}{(+3.21)}        \\
					& $\mathcal{AP}_{seg}$     & 54.37      \textcolor{blue}{(+5.56)}    & 65.04    \textcolor{blue}{(+5.98)}      & 61.48 \textcolor{blue}{(+5.84)}         \\ \hline
					\multirow{3}{*}{Pertrained} & $\mathcal{AUROC}_{det}$ & \textbf{99.70} \textcolor{blue}{(+3.14)}& \textbf{98.36} \textcolor{blue}{(+3.56)}& \textbf{98.81} \textcolor{blue}{(+3.42)} \\
					& $\mathcal{AUROC}_{seg}$   &97.80  \textcolor{blue}{(-0.06)} & \textbf{98.47} \textcolor{blue}{(+6.69)} & \textbf{98.25} \textcolor{blue}{(+4.44)} \\
					& $\mathcal{AP}_{seg}$     & \textbf{55.16} \textcolor{blue}{(+6.35)} & \textbf{67.21} \textcolor{blue}{(+8.15)} & \textbf{63.19}  \textcolor{blue}{(+7.55)}\\ \cline{1-5} 
				\end{tabular}%
		} \label{tab7}
	\end{table}

The anomaly detection and anomaly segmentation results for each category are reported in Table~\ref{tab5-2}.
Image Inpainting model (ImI) shows very poor anomaly detection performance (achieved 81.33\% in terms of $\mathcal{AUROC}_{det}$ ). Since the masked region loses the information, it leads to a large error in the recovery of the normal regions, even though it obtains large error in the recovery of the abnormal regions. Especially in the recovery of detail-sensitive categories, the results of ImI are poor. For example, in the ``Screw" and ``Grid" categories, these types of anomaly regions are extremely subtle and require accurate recovery of normal areas. Additionally, IIHP addresses this problem by introducing the HOG image prompt to recover detailed information. The performance of IIHP is significantly improved compared with ImI, which validates HOG image is effective. HOG images as prompts can provide corresponding low-level information for the masked regions, and in combination with the contextual information, the recovery of detailed information can be realized. For example, as in the case of the Grid category, it is improved from 64.6\% obtained by ImI to 100\% with regard to $\mathcal{AUROC}_{det}$, and from 84.8\% obtained by ImI to 99.2\% with regard to $\mathcal{AUROC}_{seg}$. In particular, the $\mathcal{AP}_{seg}$, which is more capable of indicating the details of the anomaly segmentation, is improved from 12.55\% to 46.32\%. These significant improvements are sufficient to indicate that the HOG is sufficiently rich in providing detailed information to ensure the recovery error of the image is in line with the anomaly detection goal. As shown in Figure~\ref{fig:figure5}, the recovery results of IIHP compared to ImI, with details of the normal region in the red dashed box, are significantly better for IIHP than for ImI.
%结合图像继续描述
However, in the IIHP, the context may still contain the original anomaly information, which causes the anomaly regions to be recovered. The performance of HIP is further improved significantly compared to IIHP since it introduces normal samples as prompt images while discarding the original input. HIP avoids the interference of the presence of anomalous features in the context. As shown in Figure~\ref{fig:figure5}, the recovery result of HIP is significantly better than that of IIHP,the normal region is recovered with more details. The quantitative results of MVTec-AD are shown in Table~\ref{tab5-2}, where the overall performance of HIP is significantly better than that of IIHP. For the AP metrics, each of the majority of the subcategories demonstrates considerable performance improvement. The comprehensive analysis above clearly indicates that HOG is able to provide detailed information guidance to enhance the recovery effect. The introduction of prompts for normal images further ensures the recovery of normal regions and improves the accuracy of anomaly detection.

\begin{table}
\scriptsize
	\centering
	\renewcommand\arraystretch{1.4}
	\caption{Ablation experimental results of Recover Network on MVTec-AD~\cite{bergmann2019mvtec} and KolektorSDD2~\cite{bovzivc2021mixed}.}{
		% \resizebox{\columnwidth}{!}{%
			\setlength{\tabcolsep}{0.5mm}{
				\begin{tabular}{c|c|c|p{1.3cm}<{\centering}p{1.3cm}<{\centering}p{1.3cm}<{\centering}p{1.3cm}<{\centering}}
    % \begin{tabular}{c|c|c|cccc}
					\hline
					
					\begin{tabular}[c]{@{}c@{}}Recover\\ Net\end{tabular}                       & \multicolumn{2}{c|}{}  & \textbf{\ding{55}} & \ding{55} & \ding{52} & \ding{52} \\ \hline
					\multirow{4}{*}{\begin{tabular}[c]{@{}c@{}}Discriminate\\ Net\end{tabular}}  & \multirow{2}{*}{input} & original image  & \ding{52}  & \ding{52}  & \ding{55}  & \ding{55}  \\
					&       & recovery image & \ding{55}  & \ding{55}  & \ding{52} & \ding{52}  \\ \cline{2-7} 
					& \multirow{2}{*}{Loss}  & $L_{D}$            & \ding{52}  & \ding{52}  & \ding{52}  & \ding{52}  \\
					&       & $L_{S}$            & \ding{55} & \ding{52}  & \ding{55}  & \ding{52}  \\ \cline{1-7} 
					
					\multirow{6}{*}{MVTec-AD} & \multirow{3}{*}{Texture} &$\mathcal{AUROC}_{det}$  & 99.6     & 98.9      &  99.4       & 99.7        \\
					&                          &	$\mathcal{AUROC}_{seg}$   & 97.8     & 97.6  &  97.6           & 97.8        \\
					&                          &	$\mathcal{AP}_{seg}$   & 53.6     & 54.0      &     54.6      & \textbf{55.2}        \\ \cline{2-7} 
					& \multirow{3}{*}{Object}  &$\mathcal{AUROC}_{det}$  & 98.3     & 98.5      &     96.3    & 98.4         \\
					&                          &	$\mathcal{AUROC}_{seg}$   & 98.0     & 97.9  &        97.6     & 98.5       \\
					&                          & 	$\mathcal{AP}_{seg}$  & 63.5     & 65.0       &   65.0       & \textbf{67.2}        \\ \hline
					\multicolumn{2}{c|}{\multirow{3}{*}{KolektorSDD2}}         &$\mathcal{AUROC}_{det}$  & 94.8     & 94.2          &  94.9   & 96.2        \\
					\multicolumn{2}{c|}{}                           & $\mathcal{AUROC}_{seg}$ & 98.2     & 98.0            &  98.3   & 98.8        \\
					\multicolumn{2}{c|}{}                          & 	$\mathcal{AP}_{seg}$  & 47.7     & 49.3            &  53.1 & \textbf{54.6}        \\ \cline{1-7} 
				\end{tabular}%
	}} \label{tab4}
\end{table}
% \end{wraptable}

\subsubsection{The effects of the Recover Network} we explore the effects of introducing Recover Network in \textit{ReDi} for anomaly detection and anomaly segmentation. When the Recover Network is incorporated, the inputs to the reference and recovery branches of the Discriminant Network are the original image and the recovery image, respectively. In contrast, without the Recover Network, both branches receive the original image as input. Experiments are performed on the MVTec-AD and KolektorSDD2 datasets, and the results are reported in Table \ref{tab4}.
The introduction of the Recover Network results in performance improvement on both datasets under two different settings (with or without the incorporation of self-correlation loss). For instance, when the self-correlation loss is included, on the KolektorSDD2 dataset, the $\mathcal{AP}{seg}$ improves from 49.3\% to 54.6\%, and the AUROC for anomaly detection enhances from 94.2\% to 96.2\%. Similarly, on the MVTec-AD dataset, both the Texture"-type and Object"-type data show performance improvements. Considering the setting with self-correlation loss ($L_S$) included as an example, the average $\mathcal{AP}{seg}$ for Texture" improves from 54.0\% to 55.2\%, and for Object" it improves from 65.0\% to 67.2\%. Analogous results can be inferred in the other setting. The Recover Network serves as an anomaly filter for the recovery branch of the Discriminant Network, enabling the features extracted by the recovery branch to effectively filter out anomalies. Consequently, the disparity in features generated in the anomaly regions becomes more pronounced.

	% \begin{figure}
	% 	\centering
	% 	\includegraphics[width=0.99\linewidth]{picture/HOG_2}
	% 	\caption{HOG images of the abnormal samples. ``HOG-1" indicates $bin=9$ and the local area $size$ is $8\times 8$. ``HOG-2" indicates $bin=18$ and the local area $size$ is $4\times 4$. It can be found that the ``HOG-2" prompt information is more detailed.}
	% 	\label{fig:hog}
	% \end{figure}

 % \begin{table}[]

% \usepackage{graphicx}

\subsubsection{The effects of self-correlation loss $L_S$} First, we analyze the effectiveness of the proposed self-correlation loss $L_S$ by comparing the results of the \textit{ReDi} method with and without $L_S$ on two benchmark datasets, as shown in Table \ref{tab4}. The group with $L_S$ outperforms the group without $L_S$, indicating that the self-correlation loss is effective for anomaly detection. Second, we investigate the general applicability of $L_S$ by incorporating it into two standard distillation models, and the results are presented in Table \ref{tab5-1}. After introducing $L_S$, the anomaly detection and anomaly segmentation performance of all models are improved. For example, in the standard distillation method with WideResNet50 as the backbone, each of the three metrics is improved by about 3\% with the introduction of the self-correlation loss, which validates the powerful generality of $L_S$. The most probable reason for effectiveness is that $L_S$ further constrains the consistency of the distribution of $F_X$ and $F_Y$ by the correlation constraint. It ensures that the features extracted by the two branches for normal regions are significantly consistent, thus improving accuracy in identifying normal areas.
\begin{wrapfigure}{r}{0.5\textwidth}
\centering
\includegraphics[width=0.99\linewidth]{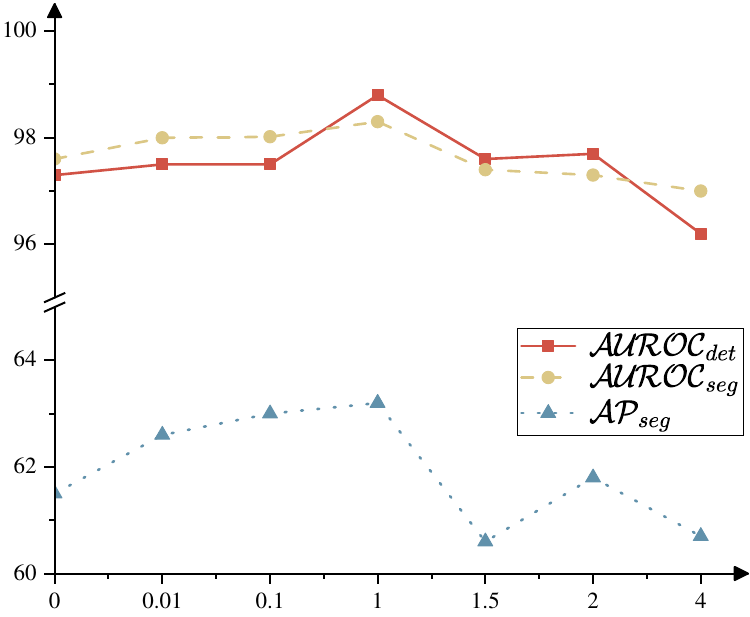}
		\caption{The effect of the weight parameter $\lambda_S$.}
  \vspace{-0.5cm}
		\label{fig:graph8}
\end{wrapfigure}

% \begin{figure}[htbp]
% \centering
% \begin{minipage}[t]{0.45\textwidth}

% \end{minipage}
% \begin{minipage}[t]{0.52\textwidth}

% \end{minipage}
% \end{figure}

\subsubsection{Prompt image $X^+$ of HIP} Self-generated maps and prompt images are important components in HIP. The results of ablation experiments (1) and (2) demonstrate the effectiveness of self-generated maps. To explore the importance of prompt image $X^+$ as another input to HIP, we set 1) no prompt image( `Free' ), 2) sampling prompt image with self-supervised method (`Self supervised'), and 3) sampling prompt image with pre-trained model (`Pertrained'), where the difference between setting 2) and 3) is different methodes of extracting features of normal image. If the prompt image is missing (`Free'), the recovery task becomes an unconditional image generation task, which is difficult to guarantee the quality of the recovery image. If the similarity between $X^+$ and $X$ is low, the prompt image cannot provide effective normal feature prompt. Table \ref{tab7} shows the experimental results of extracting $X^+$ by these three strategies to achieve HIP (including `Free'). As expected, without the prompt images, the performance of anomaly detection and anomaly segmentation is significantly reduced. The anomaly detection performance ($\mathcal{AUROC}_{det}$) decreases by 2\%-3\% and the anomaly segmentation performance ($\mathcal{AUROC}_{det}$) decreases by 5\%-8\% compared to the strategy that utilizes prompt images. Compared with the features extracted by the self-supervised model, the pre-trained model is better. $X^+$ sampled by pre-trained models improves anomaly detection and anomaly segmentation metrics by 1\%-3\% over self-supervised model. These experimental results show that the prompted images are effective.

 % --------------------------------------------------------------------------------------------------------------------------
  \begin{wraptable}{r}{0.56\textwidth}
\scriptsize
		\centering
		\renewcommand\arraystretch{1.5}
		\caption{Anomaly detection and anomaly segmentation results of HOG images extracted with different parameters serve as input of HIP on MVTec-AD. '\uline{$\mathcal{AUROC}_{det}$}, \uwave{$\mathcal{AUROC}_{seg}$}, and \textbf{$\mathcal{AP}_{seg}$}' represents the best results for $\mathcal{AUROC}_{det}$, $\mathcal{AUROC}_{seg}$, and $\mathcal{AP}_{seg}$, respectively.}
		\renewcommand\arraystretch{1.2}
		% \resizebox{\columnwidth}{!}{%
			\begin{tabular}{cc|c|ccc|c}
				\hline
				\multirow{2}{*}{$bin$} & \multirow{2}{*}{$size$} &\multirow{2}{*}{Metric} & \multicolumn{3}{c|}{MVTec-AD} & Kolek-  \\ \cline{4-6} 
				&                       &   & Texture  & Object  & MEAN   &  torSDD2     \\ \cline{1-7}
				\multirow{3}{*}{4}   & \multirow{3}{*}{$8\times 8$}    & $\mathcal{AUROC}_{det}$ & 97.46    & 97.94   & 97.78  & 94.7     \\
				&                       &	$\mathcal{AUROC}_{seg}$   & 98.46    & 96.56   & \uwave{98.60}  &98.7       \\
				&                       & $\mathcal{AP}_{seg}$  & 52.84    & 66.05   & 61.65  & 54.4      \\  \cline{1-7}
				\multirow{3}{*}{9}   & \multirow{3}{*}{$8 \times 8 $}    &$\mathcal{AUROC}_{det}$  & 99.70    & 98.36   & \uline{98.81}  & \uline{96.2}  \\
				&                       & 	$\mathcal{AUROC}_{seg}$  & 97.80   & 98.47   & 98.25  & \uwave{98.8}  \\
				&                       &$\mathcal{AP}_{seg}$   & \textbf{55.16 }   & \textbf{67.21}   & \textbf{63.19}  & \textbf{54.6} \\  \cline{1-7}
				\multirow{3}{*}{18}  & \multirow{3}{*}{$4 \times 4$}    &$\mathcal{AUROC}_{det}$  & 97.04    & 97.98   & 97.67  & 94.8  \\
				&                       & 	$\mathcal{AUROC}_{seg}$  & 99.02    & 96.82   & 97.55  & 98.7  \\
				&                       &$\mathcal{AP}_{seg}$   & 54.70    & 65.81   & 62.11  & 53.4 \\ \cline{1-7} 
			\end{tabular}\label{tab8}
		
% \end{table}
\end{wraptable}
\subsubsection{The effects of different HOG images $H$ of HIP} 
When extracting HOG, different hyperparameters may bring HOG images with different gradient details. As shown in Figure \ref{fig:hog}, HOG images with different hyperparameters (orientation ($bin$) and local region size ($size$)) are demonstrated. As $bin$ increases and $size$ decreases, the HOG image is able to describe more details of the image. However, this may be able to pose the risk of exposing anomalies. To explore the effect of different hyperparameters on the performance of HIP, we extract three sets of HOG images with different parameters as inputs to HIP and conducted experiments on MVTec-AD and KolektorSDD2, respectively. The comprehensive results are shown in Table \ref{tab8}. For $\mathcal{AUROC}_{det}$ and $\mathcal{AP}_{seg}$, the best performance is achieved when $bin=9$ and $size$ is $8\times 8$. When the low-level information indicated by the HOG image is not enough to show the details of the normal region of the image (e.g., $bin=4$ and $size$ is $8\times 8$), the recovery results of HIP are poor, resulting in low accuracy of anomaly detection and anomaly segmentation. For the two publicly available anomaly detection datasets, the optimal anomaly detection and anomaly segmentation results are obtained for $bin=9$ and $size$ of $8\times 8$.

% The larger the $bin$ and the smaller the $size$, the more detailed information is expressed by HOG. The comprehensive results reported in Table \ref{tab8} show the best performance at $bin=9$, size of $8\times 8$. The reason may be that the excessively fine feature information can instead support the recovery of anomalous regions. Too coarse-grained features cannot support the recovery of normal regions.  $bin=9$ and $size = 8\times 8$ better trade-offs the recovery of abnormal and normal regions. 

% --------------------------------------------------------------------------------------------------------------------------
\subsubsection{The effect of the weight parameter $\lambda_S$ for self-correlation loss}
{
In Discriminant Net, $\lambda_S$ affects the weights of two losses. 
To explore this, we conduct experiments where $\lambda_S$ is varied over a broader range, with detailed results now presented in Figure~\ref{fig:graph8}. These experiments were designed to isolate the effects of $\lambda_S$, keeping other parameters constant, to ensure a clear view of its impact on ReDi’s effectiveness. Our findings indicate that the performance of ReDi in terms of anomaly detection and localization improves as the weight of self-correlation loss increases, reaching an optimal balance with the raw cosine similarity loss at $\lambda_S =1$. Beyond this value, we observed a decrease in performance. We theorize that this decline is due to an overemphasis on self-correlation, which may overshadow the contribution of other vital features, leading to less effective anomaly discrimination.
% 
% We explore the impact of $\lambda_S$ on the anomaly detection and anomaly segmentation performance of \textit{ReDi}. As shown in Figure \ref{fig:graph8}, the results demonstrate that the anomaly detection performance improves as the $L_S$ constraint becomes stronger. The optimal setting of $\lambda$ is $\lambda =1 $.
}

	% 	\begin{figure}
	% 	\centering
	% 	\includegraphics[width=1\linewidth]{picture/Graph_7_2.eps-60305}
	% 	\caption{The effect of the weight parameter $\lambda_S$ for self-correlation loss.}
	% 	\label{fig:graph8}
	% \end{figure}

% 	\begin{table}[]
% \tiny
% \color{red}
% 	\centering
% 	\renewcommand\arraystretch{1.3}
% 	\caption{{Anomaly detection and anomaly segmentation results of Feature Reconstruction Block on  MVTec-AD. `FRB' represents the Feature Reconstruction Block.}}
% 	\renewcommand\arraystretch{1.5}
% 	% \resizebox{\columnwidth}{!}{%
% 		\begin{tabular}{@{}cc|lll@{}}
% 			\toprule
% 			 HIP & FRB &  $\mathcal{AUROC}_{det}$     & $\mathcal{AUROC}_{seg}$      & $\mathcal{AP}_{seg}$      \\ \hline
% 			  \ding{55}   & \ding{55}    &97.61&96.67&58.11\\
%  \hdashline
% 			% \ding{52}    & \ding{55}    & \ding{55}    & 96.01\textcolor{blue}{\tiny{(-1.60)}} & 97.03\textcolor{blue}{\tiny{(+0.36)}} & 59.61\textcolor{blue}{\tiny{(+1.50)}} \\
%         \ding{52}   & \ding{55} &97.23\textcolor{blue}{\tiny{(-0.38)}}&97.09\textcolor{blue}{\tiny{(+0.42)}}&62.33\textcolor{blue}{\tiny{(+4.22)}}\\
% 			 \ding{52}   & \ding{52}   & 98.81\textcolor{blue}{\tiny{(+1.20)}} & 98.25\textcolor{blue}{\tiny{(+1.58)}} & 63.19\textcolor{blue}{\tiny{(+5.08)}} \\ \hline
% 		\end{tabular}%
% 	 \label{fig9}
% \end{table}

\begin{table}
\scriptsize
\parbox{.49\linewidth}{
% \color{red}
	\centering
	\renewcommand\arraystretch{1.3}
	\caption{{Anomaly detection and anomaly segmentation results of Feature Reconstruction Block on  MVTec-AD. `FRB' represents the Feature Reconstruction Block.}}
	\renewcommand\arraystretch{1.5}
	% \resizebox{\columnwidth}{!}{%
		\begin{tabular}{@{}cc|lll@{}}
			\toprule
			 HIP & FRB &  $\mathcal{AUROC}_{det}$     & $\mathcal{AUROC}_{seg}$      & $\mathcal{AP}_{seg}$      \\ \hline
			  \ding{55}   & \ding{55}    &97.61&96.67&58.11\\
 \hdashline
			% \ding{52}    & \ding{55}    & \ding{55}    & 96.01\textcolor{blue}{\tiny{(-1.60)}} & 97.03\textcolor{blue}{\tiny{(+0.36)}} & 59.61\textcolor{blue}{\tiny{(+1.50)}} \\
        \ding{52}   & \ding{55} &97.23\textcolor{blue}{\tiny{(-0.38)}}&97.09\textcolor{blue}{\tiny{(+0.42)}}&62.33\textcolor{blue}{\tiny{(+4.22)}}\\
			 \ding{52}   & \ding{52}   & 98.81\textcolor{blue}{\tiny{(+1.20)}} & 98.25\textcolor{blue}{\tiny{(+1.58)}} & 63.19\textcolor{blue}{\tiny{(+5.08)}} \\ \hline
		\end{tabular}%
	 \label{fig9}
}
\hfill
\parbox{.49\linewidth}{
% \color{red}
\centering
\caption{Comparison of complexity (parameters (M), FLOPs (G)) and inference time between our \textit{ReDi} and state-of-the-art methods. }{
	\begin{tabular}{cccc}	
		\hline
Method        & Parameters(M) & FLOPs(G) & Inference Time(s) \\ \hline
RIAD~\cite{zavrtanik2021reconstruction}  & 28.8          & 222.1  & 0.2234            \\
MKDAD~\cite{salehi2021multiresolution} & 138.4         & 30.8   & 0.0858            \\
RD~\cite{deng2022anomaly}    & 83.8          & 38.9   & 0.0715            \\
RD++~\cite{Tien_2023_CVPR}  & 166.1         & 62.0   & 0.1254            \\
ReDi          & 106.4         & 107.2  & 0.1086            \\ \hline
\end{tabular}\label{tabsupp2}
}} 
\end{table}

% --------------------------------------------------------------------------------------------------------------------------
\subsubsection{The effect of Feature Recovery Block} This study explores the effect of Feature Recovery Block of the $ReDi$ framework. To ensure that the recovery branch is able to learn features in the normal regions consistent with the reference branch, Extractor $3$ is changed to a learnable encoder when the FRB is removed. Extractor $3$ is served as the recovery feature. The ablation experimental results are shown in Table \ref{fig9}.
It can be found that the anomaly detection performance can be significantly improved by using the FRB in combination with HIP compared to the original baseline.FRB can further improve the results after the introduction of HIP, e.g., the AP is improved from 62.33\% to 63.19\%. FRB ensures that the normal features are aligned while the anomalous features are further filtered. Therefore, \textit{ReDi} possesses powerful anomaly detection and anomaly segmentation capabilities.

\begin{figure*}

    \centering
    \includegraphics[width=0.99\linewidth]{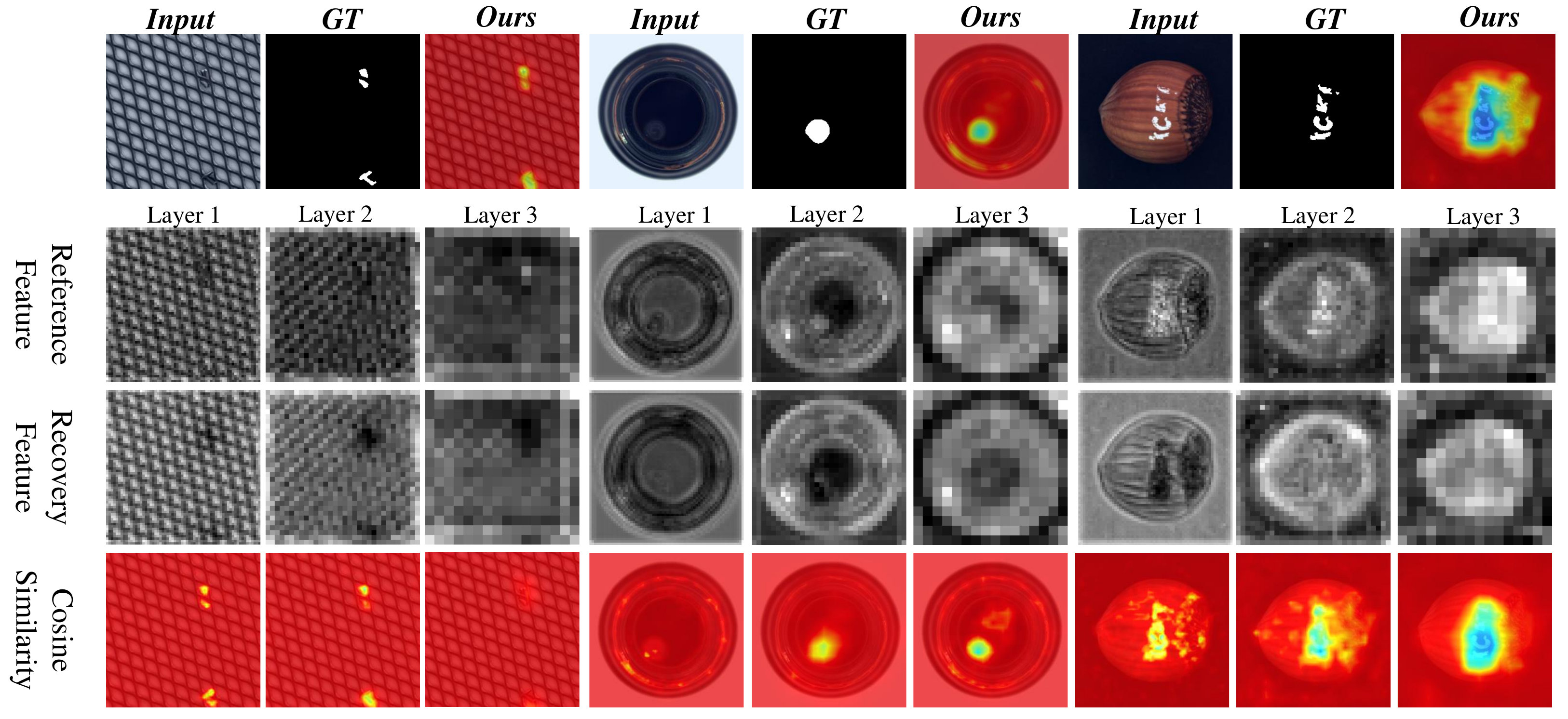}
    \caption{\color{red}{Visualization results for the reference features and recovery features.}}
    \label{fig:sup1}
\end{figure*}

{
Figure~\ref{fig:sup1} clearly demonstrates the difference between the reference and recovery features in the anomalous regions. In addition, we also observe differences in the detection of anomalous regions in different layers due to differences in the receptive fields. For example, in the first anomaly example, the anomalous region can be detected in both ``layer 1'' and ``layer 2'', but it is not obvious in ``layer 3''. In the second anomaly example, the layer1 layer fails to detect the anomalous region significantly. Therefore, we use the weighted anomaly detection results of the three layers to obtain the final anomaly localization result ("Ours".)}

\subsubsection{Complexity and Inference Time}
{
We include in Table~\ref{tabsupp2} a comprehensive comparison of the proposed \textit{ReDi} against recent anomaly detection methods~\cite{zavrtanik2021reconstruction,deng2022anomaly,Tien_2023_CVPR,salehi2021multiresolution}, focusing on model complexity (as measured by the number of parameters and FLOPs) and inference time. The \textit{ReDi} framework, through its innovative HIP recovery process, exhibits a slightly longer inference time compared to the direct distillation scheme RD~\cite{deng2022anomaly}. This increase is primarily due to the additional step of recovering the HOG image, which is integral to our method's enhanced detection capability. Despite this, it's important to note that our approach remains highly competitive in terms of efficiency. For instance, compared to the RIAD~\cite{zavrtanik2021reconstruction}, which relies on an inpainting technique requiring multiple forward processes, our \textit{ReDi} achieves faster inference times with only a single forward pass. Integrating the results in Table~\ref{TAB.2} of the original manuscript, our method better weighs the parameters and inference time.}

\begin{wrapfigure}{r}{0.4\textwidth}
\vspace{-1.4cm}
\centering
\includegraphics[width=0.99\linewidth]{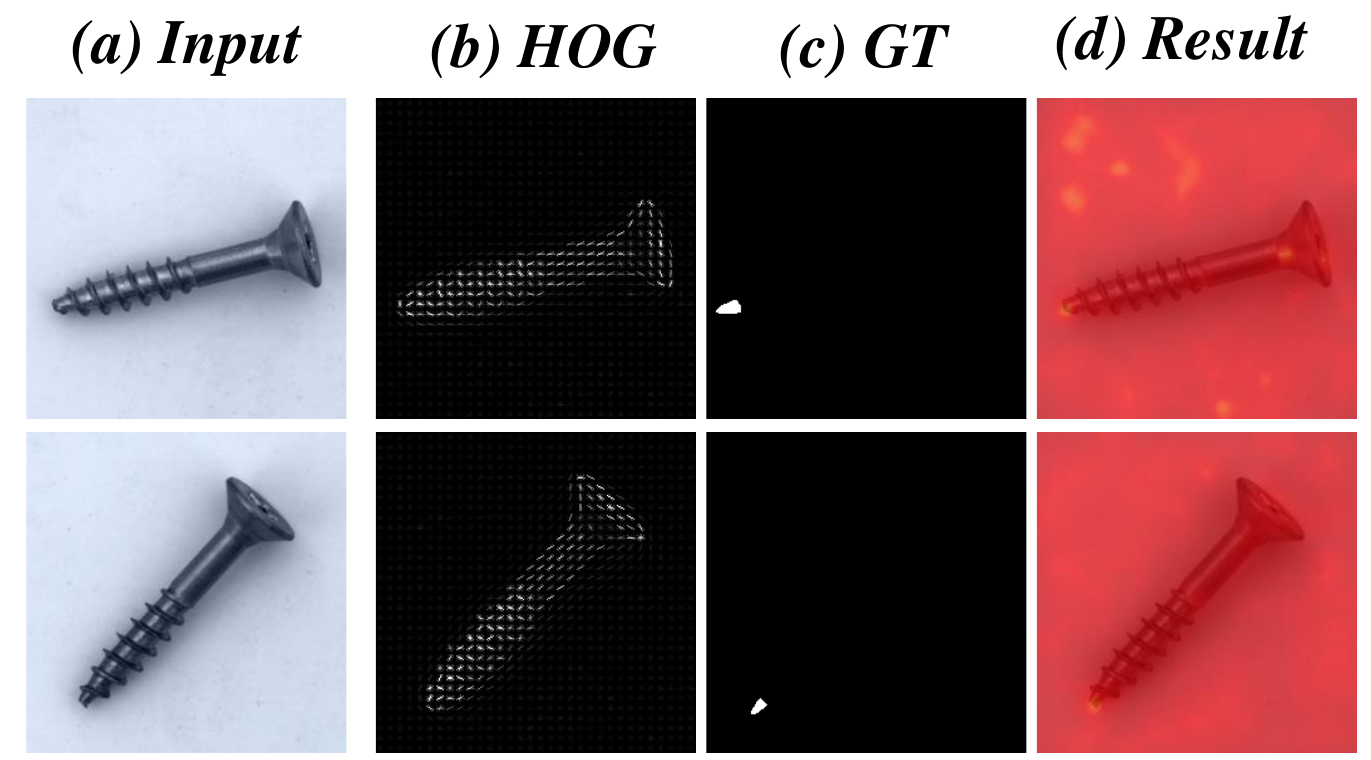}
		\caption{Failure cases of the proposed \textit{ReDi}.}
        \vspace{-0.4cm}
		\label{fig:lim}
\end{wrapfigure}

\subsection{Limitation}

	% \begin{figure}
	% 	\centering
	% 	\includegraphics[width=0.99\linewidth]{picture/limi}
	% 	\caption{Failure cases of the proposed \textit{ReDi}.}
 %        \vspace{-2mm}
	% 	\label{fig:lim}
	% \end{figure}

Although the proposed method effectively addresses most challenges associated with anomaly detection and segmentation, its performance is somewhat compromised when dealing with 'missing' types of anomalies that are not indicated by the Histogram of Oriented Gradients (HOG) map. For instance, as depicted in Figure \ref{fig:lim}, our approach, \textit{ReDi}, shows limited success in the 'screw' class where the subject is missing the tip part. HOG struggles to capture the complete low-level information of this region, which subsequently leads to an inadequate recovery by the  HIP, and thus \textit{ReDi} fails to segment the abnormal region. In future work, we will improve our method by introducing low-level information guidance of normal images.

%------------------------------------------------------------------------

% ----------------------------------------
%------------------------------------------------------------------------
\section{Conclusion}\label{sec5}
In this paper, we proposed an anomaly detection framework called \textit{ReDi}, which comprised of a Recover Network and a Discriminate Network. In the Recover Network, we proposed a recovery method named HIP to address two potential problems in anomaly detection. The HIP method incorporated both structural and normal semantic information to ensure the recovery of normal regions while avoiding the capture of original anomalies. Furthermore, in the Discriminate Network, we segmented anomalous regions by identifying differences in the feature space with the aid of a pre-trained model. We conducted extensive experiments to validate the effectiveness of each module. In the future, we aim to develop more advanced prompt methods to further enhance the anomaly detection performance.

% \newpage

% \footnotesize

%%%%%%%%%%%%%%%%%%%%%%%%%%%%%%%%%%%%%%%%%%%%%%%%%%%%%%%
%%% Acknowledgements. ÖÂÐ»
%%%%%%%%%%%%%%%%%%%%%%%%%%%%%%%%%%%%%%%%%%%%%%%%%%%%%%%
\Acknowledgements{This work was supported by the National Natural Science Foundation of China (Grant No. U21B2043).  The authors would like to thank all the anonymous reviewers for their constructive comments and suggestions.}

%%%%%%%%%%%%%%%%%%%%%%%%%%%%%%%%%%%%%%%%%%%%%%%%%%%%%%%
%%% Supplements. ²¹³ä²ÄÁÏ, ·Ç±ØÑ¡
%%%%%%%%%%%%%%%%%%%%%%%%%%%%%%%%%%%%%%%%%%%%%%%%%%%%%%%
% \Supplements{Appendix A.}

%%%%%%%%%%%%%%%%%%%%%%%%%%%%%%%%%%%%%%%%%%%%%%%%%%%%%%%
%%% Reference section. ²Î¿¼ÎÄÏ×
%%% citation in the content using "some words~\cite{1,2}".
%%% ~ is needed to make the reference number is on the same line with the word before it.
%%%%%%%%%%%%%%%%%%%%%%%%%%%%%%%%%%%%%%%%%%%%%%%%%%%%%%%
{
\bibliographystyle{bib}
\bibliography{egbib}
}
% \begin{thebibliography}{99}

% \end{thebibliography}

%%%%%%%%%%%%%%%%%%%%%%%%%%%%%%%%%%%%%%%%%%%%%%%%%%%%%%%
%%% Appendix sections. ¸½Â¼ÕÂ½Ú, ·Ç±ØÑ¡
%%%%%%%%%%%%%%%%%%%%%%%%%%%%%%%%%%%%%%%%%%%%%%%%%%%%%%%
%\begin{appendix}
%\section{Name}

%\end{appendix}

\end{document}